%% file: iclr2027_conference.tex
\documentclass{article} % For LaTeX2e
\PassOptionsToPackage{dvipsnames}{xcolor}
\usepackage{iclr2027_conference,times}

\input{math_commands.tex}

\usepackage{hyperref}
\hypersetup{pdftitle={Let Androids Dream: From Visual Elements to Emergent Implications via Contextual Alignment},pdfauthor={Chenhao Zhang, Yazhe Niu}}
\usepackage{url}

\usepackage[utf8]{inputenc}
\usepackage[T1]{fontenc}
\usepackage[dvipsnames]{xcolor}
\usepackage{booktabs}
\usepackage{nicefrac}
\usepackage{microtype}
\usepackage{inconsolata}
\usepackage{CJKutf8}
\usepackage{pifont}
\usepackage{graphicx}
\usepackage[caption=false]{subfig}
\usepackage{multirow}
\usepackage{adjustbox}
\usepackage{soul}
\usepackage{enumitem}
\usepackage{wrapfig}
\usepackage[linesnumbered,ruled,vlined]{algorithm2e}
\usepackage{float}
\usepackage{listings}
\usepackage{colortbl}

\definecolor{darkblue}{rgb}{0, 0, 0.5}
\definecolor{darkred}{RGB}{255, 150, 113}
\definecolor{darkgreen}{RGB}{50, 206, 50}

\long\def\rev#1{#1}

\newcommand{\up}{\textcolor{darkred}{$\uparrow$}}
\newcommand{\down}{\textcolor{darkgreen}{$\downarrow$}}

\hypersetup{colorlinks=true, citecolor=darkblue, linkcolor=darkblue, urlcolor=darkblue}
\newcommand{\ablationindent}{\hspace*{1.5em}}

\setlist[itemize]{leftmargin=*, topsep=1pt, itemsep=0.5pt, parsep=0pt, partopsep=0pt}
\setlist[enumerate]{topsep=1pt, itemsep=0.5pt, parsep=0pt, partopsep=0pt}

\newcommand{\MLLM}{\text{MLLM}}

\newcommand{\WebSearchStratText}{\text{`WebSearch'}}

\title{Let Androids Dream: From Visual Elements to Emergent Implications via Contextual Alignment}

\author{Chenhao Zhang \\
Shanghai AI Laboratory \\
Huazhong University of Science and Technology \\
\texttt{zhangchenhao@pjlab.org.cn} \\
\And
Yazhe Niu \\
Shanghai AI Laboratory \\
The Chinese University of Hong Kong \\
\texttt{niuyazhe@pjlab.org.cn}}

\iclrfinalcopy
\begin{document}

\maketitle
% Running head for the preprint.
\lhead{Preprint}

\begin{abstract}
Image metaphor understanding differs from general VQA: the implied meaning is rarely explicit in pixels and instead emerges from the \emph{compositional interaction} of visual elements together with social intent and culture-specific background. \rev{We identify \emph{missing context}, namely cultural, social, or rhetorical conventions already in a model's weights that single-pass inference leaves unused and external facts absent from the weights, as a failure mode of current multimodal large language models (MLLMs).} We formalize \emph{Contextual Alignment} as inference over a latent context: selecting background knowledge that is relevant to the image's composition and useful for inferring its implication. We then propose \textbf{\textit{Let Androids Dream (LAD)}}, a training-free framework that operationalizes contextual alignment through test-time computation: (1) \textbf{Perception} extracts compositional visual structure, (2) \textbf{Search} \rev{adaptively acquires context from parametric knowledge or web search}, and (3) \textbf{Reasoning} aligns compositions with context to derive emergent implications. On S100, a benchmark of 100 high-level implication questions in English and Chinese, LAD raises the lightweight GPT-4o-mini from 44\% to 74\% (English) and from 42\% to 52\% (Chinese) Multiple-Choice Question (MCQ) accuracy, on par with the closed-book GPT-5.4 in English, and from 2.98 to 4.02 and from 3.36 to 3.66 on Open-Style Questions (OSQ). LAD improves all \rev{five} backbones we test, including the open-source Qwen2.5-VL-7B (+18 MCQ and +1.30 OSQ points in English)\rev{, GPT-4o (4.14 and 4.26 OSQ), and Gemini-3.0-Pro, where it obtains the highest OSQ scores (4.30 and 4.46)}, and its gains hold on the full II-Bench and CII-Bench. Adding other search tools raises MCQ accuracy in five of six cases but changes OSQ scores by $-0.48$ to $+0.01$, whereas LAD's search stage improves both, suggesting that how context is acquired matters beyond access to search. Our OSQ judge is within one point of trimmed human means for 95.7\% of answers, and LAD with web search also improves MMMU, SeedBench, and MMStar. We open-sourced all code and datasets at \url{https://ming-zch.github.io/lad}.
\end{abstract}

\input{sec/intro}

\input{sec/related_work}

\input{sec/method}

\input{sec/experiments}

\input{sec/discussion}

\input{sec/conclusion}

\clearpage
\bibliography{custom}
\bibliographystyle{iclr2027_conference}

\newpage

\appendix

\input{sec/appendix}

\end{document}

%% file: math_commands.tex
\usepackage{amsmath,amsfonts,bm}

\def\eqref#1{equation~\ref{#1}}
\def\1{\bm{1}}

\DeclareMathAlphabet{\mathsfit}{\encodingdefault}{\sfdefault}{m}{sl}
\SetMathAlphabet{\mathsfit}{bold}{\encodingdefault}{\sfdefault}{bx}{n}

%% file: sec/intro.tex
\section{Introduction}
\label{sec:intro}

\begin{quote}\textit{Do androids dream of electric sheep? The question actually has two levels: The first level is to ask if androids dream, and the second level is to ask if they dream of electric sheep.}\par\raggedleft -- Philip K. Dick (1968)\end{quote}

Metaphors pervade daily communication -- from ``time is money'' to visual satire in editorial cartoons, serving as fundamental cognitive tools for conceptualizing abstract ideas \citep{lakoff2008metaphors}.
Enabling AI to understand such metaphors, particularly in the visual modality, requires going far beyond literal object recognition, and it matters in practice for moderating harmful memes, analysing political communication, and interpreting images across cultures.
Critically, the meaning of an image metaphor does not reside in individual visual entities themselves, but emerges from their \emph{compositional interaction} under contextual and cultural priors -- making this task qualitatively different from standard visual question answering.

\begin{wrapfigure}{r}{0.45\textwidth}
    \centering
    \vspace{-10pt}
    \includegraphics[width=\linewidth]{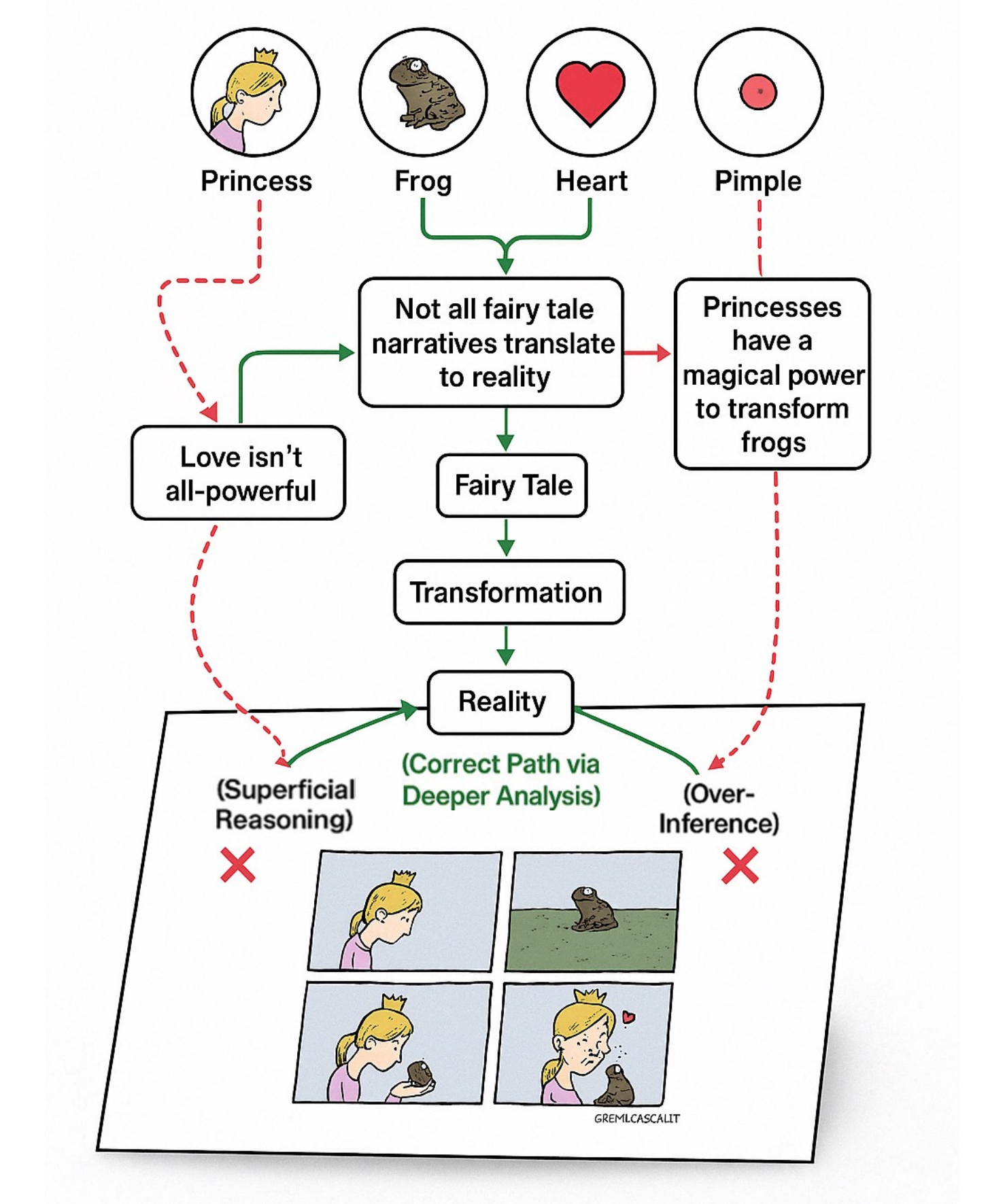}
    \caption{An image is worth a thousand words: For image metaphor understanding, different element combinations lead to different reasoning paths, while the correct path requires integrating all elements.}
    \label{fig:teaser}
    \vspace{-8pt}
\end{wrapfigure}

We identify a systematic failure mode in current multimodal reasoning: \emph{missing context}. \rev{Interpreting image metaphors requires connecting compositional visual structures to cultural, social, or rhetorical background (Figure~\ref{fig:teaser}), which may be either conventions already stored in the model's weights that single-pass inference leaves unused or external facts genuinely absent from the weights.}
Existing approaches either attempt explicit metaphor mapping \citep{zhong2024lets}, which is brittle under many-to-many relations, or \rev{unguided} chain-of-thought prompting \citep{xu-etal-2024-exploring}, which \rev{neither reliably elicits composition-relevant conventions from the weights nor accesses external facts}.
Scaling model parameters or task-specific training, including end-to-end reinforcement learning \citep{zhang2026metaphorstar}, can partially internalize such knowledge through train-time investment \citep{liu2024iibenchimageimplicationunderstanding,zhang2024mllmsunderstanddeepimplication}, but these approaches are costly, opaque, and model-bound. A fundamental question remains: can we achieve effective contextual alignment at inference time, transparently and without any training?

We answer this affirmatively. We first formalize \emph{Contextual Alignment} with a latent-context decomposition: the implication of an image depends on background knowledge that is not visible in it, and contextual alignment selects the context that is relevant to the image's composition and useful for inferring its implication (Section~\ref{sec:formulation}). We then propose \textbf{\textit{Let Androids Dream (LAD)}}, a training-free framework that operationalizes contextual alignment through three complementary stages: \textbf{Perception} to extract compositional visual structure, \textbf{Search} to \rev{adaptively acquire missing context from parametric knowledge or web search} guided by perceived compositions, and \textbf{Reasoning} to align compositions with acquired context for implication inference.
Unlike \rev{unguided} chain-of-thought prompting \rev{(which lacks compositional perception and external access)}, vanilla retrieval augmentation (which retrieves without perception guidance), or end-to-end training (which requires costly optimization), LAD operates at inference time on a frozen MLLM and is backbone-agnostic by design. Its technical novelty lies in three components that generic retrieval pipelines lack (Table~\ref{tab:positioning}): queries are planned from the \emph{composition} of the image -- keyword combinations naming its emotion, domain, and rhetoric -- rather than from entity names; retrieved answers are ranked and refined by their relevance to the image's \emph{implication} rather than to a question; and, unlike end-to-end and RL-trained methods, the acquired context is an explicit variable that can be inspected, removed, or replaced.

On S100, a set of 100 high-level implication questions from II-Bench and CII-Bench, LAD raises the multiple-choice (MCQ) accuracy of the lightweight GPT-4o-mini from 44\% to 74\% (English) and from 42\% to 52\% (Chinese), on par with the closed-book GPT-5.4 in English, and its open-style (OSQ) score from 2.98 to 4.02 and from 3.36 to 3.66; \rev{and LAD obtains the highest OSQ scores with Gemini-3.0-Pro (4.30 and 4.46)}. Because LAD uses search, we also compare it with search-augmented systems: other search tools raise MCQ accuracy in five of six cases but change OSQ by $-0.48$ to $+0.01$, whereas LAD improves both, and LAD improves all \rev{five} backbones we test. For context-dependent reasoning, \emph{how} inference-time computation acquires context matters, not only whether the model can search.

\textit{Our key contributions are as follows:}
\begin{itemize}[leftmargin=*, noitemsep]
\item \textbf{Problem Reformulation:} We reformulate image metaphor understanding as a context-dependent compositional reasoning problem and formalize it as inference over a latent context, which makes missing context explicit and maps perception, context acquisition, and reasoning to separable stages.
\item \textbf{Failure Mode Identification:} We identify \emph{missing context} as a failure mode of frozen MLLMs; ablations show that explicit chain-of-thought alone helps little, whereas structured perception with explicit reasoning (Stage I+III) gives most of the gain on three of four metrics and perception-guided context acquisition (Stage II) adds the rest.
\item \textbf{Test-Time Contextual Alignment:} We propose LAD, a training-free, backbone-agnostic framework that achieves contextual alignment through test-time computation, and compare it with both closed-book and search-augmented systems. We also design an Open-Style Question (OSQ) evaluation protocol whose automatic scores are within one point of human ratings for 95.7\% of answers.
\item \textbf{Generalizability:} LAD improves \rev{five} backbones (Qwen2.5-VL-7B/72B, GPT-4o-mini, GPT-4o\rev{, Gemini-3.0-Pro}) on both languages and formats, the GPT-4o-mini gain holds on the 2,199 full-benchmark items, and LAD with web search improves MMMU, SeedBench, and MMStar.
\end{itemize}

%% file: sec/related_work.tex
\section{Related Work}
\label{sec:relatedWork}

\begin{table}[b]
\centering
\footnotesize
\setlength{\tabcolsep}{6pt}
\caption{Positioning of LAD. \emph{Live retrieval}: web search at test time (Self-RAG retrieves from a static index); \emph{composition-guided queries}: retrieval queries derived from the combination of visual elements and their emotion, domain, and rhetoric; \emph{implication-aware selection}: retrieved content ranked and refined by relevance to the image's implied meaning; --: not applicable (no retrieval stage). LAD's WebSearch builds on MindSearch.}
\label{tab:positioning}
\begin{adjustbox}{max width=\textwidth}
\begin{tabular}{lcccccc}
\toprule
\textbf{Method} & \textbf{\shortstack{Training-\\free}} & \textbf{\shortstack{Multi-\\modal}} & \textbf{\shortstack{Live\\retrieval}} & \textbf{\shortstack{Adaptive\\retrieval}} & \textbf{\shortstack{Composition-\\guided queries}} & \textbf{\shortstack{Implication-\\aware selection}} \\
\midrule
C4MMD (CoT distillation)~\citep{xu-etal-2024-exploring} & \ding{55} & \ding{51} & \ding{55} & -- & -- & -- \\
Socratic Models~\citep{zeng2022socratic} & \ding{51} & \ding{51} & \ding{55} & -- & -- & -- \\
Self-RAG~\citep{asai2024selfrag} & \ding{55} & \ding{55} & \ding{55} & \ding{51} & \ding{55} & \ding{55} \\
MindSearch~\citep{chen2024mindsearch} & \ding{51} & \ding{55} & \ding{51} & \ding{55} & \ding{55} & \ding{55} \\
MMSearch-R1~\citep{wu2025mmsearchr1} & \ding{55} & \ding{51} & \ding{51} & \ding{51} & \ding{55} & \ding{55} \\
MetaphorStar~\citep{zhang2026metaphorstar} & \ding{55} & \ding{51} & \ding{55} & -- & -- & -- \\
\rowcolor{gray!12} LAD (ours) & \ding{51} & \ding{51} & \ding{51} & \ding{51} & \ding{51} & \ding{51} \\
\bottomrule
\end{tabular}
\end{adjustbox}
\end{table}

\subsection{Image Metaphor}

Early work on image metaphor targeted specific phenomena such as humor \citep{hessel-etal-2023-androids,horvitz-etal-2024-getting} and sarcasm \citep{Desai_Chakraborty_Akhtar_2022}. Explaining New Yorker cartoons \citep{hessel-etal-2023-androids} and Chinese pun rebus art \citep{zhang2024punrebus} requires knowledge that is not in the picture, but these studies do not supply it. Benchmarks for MLLMs followed: DeepEval \citep{yang-etal-2024-large} proposes a taxonomy, and II-Bench \citep{liu2024iibenchimageimplicationunderstanding} and CII-Bench \citep{zhang2024mllmsunderstanddeepimplication} evaluate English and Chinese images. Methods fall into three groups: explicit metaphor mapping (CLOT \citep{zhong2024lets}), which scales poorly to many-to-many mappings and new cultural references; chain-of-thought prompting (C4MMD \citep{xu-etal-2024-exploring}), which reasons only over parametric knowledge; and end-to-end reinforcement learning (MetaphorStar \citep{zhang2026metaphorstar}), which stores implication reasoning in the weights. LAD treats the task as \emph{contextual alignment} at test time and keeps the acquired context as an explicit variable.

\subsection{Vision-language Reasoning and Retrieval}

Reasoning models such as o1 \citep{o1}, DeepSeek-R1 \citep{deepseek-R1}, and QwQ \citep{qwq-32b-preview,qwen2} motivated vision-language reasoning, which converts images into textual grounding (Socratic Models \citep{zeng2022socratic}, LLaVA-CoT \citep{xu2024llavacotletvisionlanguage}, Mulberry \citep{yao2024mulberryempoweringmllmo1like}), elicits cues through multi-turn interaction (VoCoT \citep{li2024vocotunleashingvisuallygrounded}, V* \citep{vstar}), or reasons with visual intermediates (Visual Sketchpad \citep{hu2024visual}, Whiteboard-of-Thought \citep{menon2024whiteboard}, CoVT \citep{qin2025covt}); these methods need task-specific training or stay within parametric knowledge. Retrieval supplies missing knowledge through adaptive retrieval (Self-RAG \citep{asai2024selfrag}), search agents \citep{chen2024mindsearch}, and trained multimodal search agents (MMSearch-R1 \citep{wu2025mmsearchr1}, WebWatcher \citep{geng2025webwatcher}; MMSearch-Plus \citep{tao2025mmsearchplus}) that look up facts about depicted entities. \rev{In cross-modal retrieval, PVSE \citep{song2019polysemous} learns $K$ embeddings per instance at training time to handle polysemy without external knowledge.} The meaning of a metaphor, however, emerges from a \emph{composition} of elements\rev{, and requires unseen cultural conventions to infer one intended implication}, so LAD plans queries from perceived compositions and selects context by implication (Table~\ref{tab:positioning}).

%% file: sec/method.tex
\section{Method}
\label{sec:method}

\subsection{Problem Formulation: Contextual Alignment}
\label{sec:formulation}

Let $x$ be an image and $y$ its implication. The implication depends on background knowledge $c$ -- cultural conventions, social roles, events, idioms, puns -- that is generally not visible in $x$:
\begin{equation}
p(y\mid x)=\textstyle\sum_{c} p(y\mid x,c)\,p(c\mid x).
\label{eq:latent}
\end{equation}
An end-to-end MLLM must both read $x$ and recall a suitable $c$ from its parameters. We speak of \emph{missing context} when the model's implicit estimate of $p(c\mid x)$ misses the context $c^\star$ that the implication requires, even though the model could identify $y$ given $c^\star$. Because the relevant context depends on how elements are \emph{composed}, we estimate it from a perceptual abstraction $z=\phi(x)$ and define \emph{contextual alignment} as selecting a context $\hat c$ that is relevant to $z$ and useful for inferring $y$:
\begin{equation}
\hat c = S\big(R(Q(z))\big),\qquad \hat y=\arg\max_y\, p_\theta(y\mid x, z, \hat c),
\label{eq:lad}
\end{equation}
where $Q$ plans questions from $z$, $R$ answers them from parametric or external sources, and $S$ selects and condenses the answers. Eq.~\ref{eq:latent} is a conceptual decomposition rather than an estimable model: LAD directly constructs $\hat c$ and does not explicitly estimate either probability. Since $z$ is computed from $x$, its role is computational, and Eq.~\ref{eq:lad} exposes stages that our ablations remove or replace.

\begin{figure}[t]
    \centering
	\includegraphics[width=0.66\textwidth]{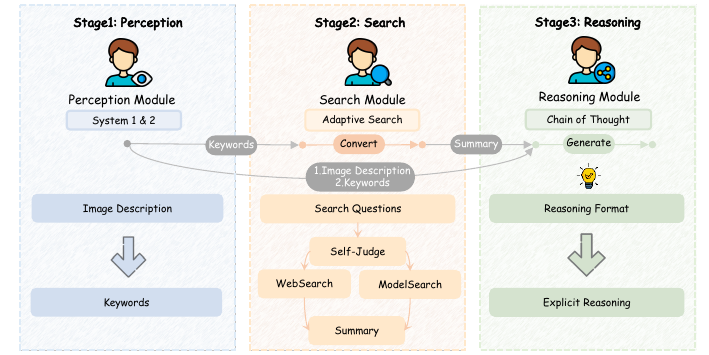}
    \caption{The general framework of Let Androids Dream (LAD), which includes three stages: (1) Perception: converting raw visual information into rich and multi-level textual representations ($z$), (2) Search: planning, routing, and condensing knowledge retrieval into a context summary ($\hat c$), and (3) Reasoning: generating context-aligned image metaphor interpretations via explicit reasoning.}
    \label{fig:method}
\end{figure}

We propose Let Androids Dream (LAD), a training-free framework that implements Eq.~\ref{eq:lad} with prompted calls to one frozen MLLM in three stages (Figure~\ref{fig:method}; Algorithm~\ref{alg:pipeline}). The same prompt templates (with Chinese versions for Chinese images) are reused for every backbone and benchmark; the English prompts are reproduced verbatim in Appendix~\ref{app:appendix_prompts}.

\subsection{Stage I: Perception}
\label{sec:perception}

This stage addresses \emph{perception failure} -- the tendency of end-to-end models to process visual elements in isolation without explicitly capturing their compositional structure.
The \textit{Perception} stage transforms raw visual inputs into structured, hierarchical textual representations ($z$). First, the MLLM produces a one-paragraph description capturing coarse-grained visual information, including discernible text within the image, prominent colors, overall layout, and salient objects or entities.
Following this, the MLLM condenses the description into seven keywords that encapsulate aspects relevant to metaphor understanding: the perceived emotion, the domain or context (e.g., political, social, cultural), and rhetorical devices, explicitly checking for puns and homophones. The keywords also re-emphasize key textual elements or entities and serve as the cues that guide knowledge retrieval in Stage~II.

\subsection{Stage II: Search}
\label{sec:search}

\rev{While Stage~I's abstraction $z$ helps Stage~III recall conventions already in the model's weights, Stage~II addresses remaining \emph{missing context}: conventions in the weights that single-pass inference still leaves unused and external facts absent from the weights, making web search adaptive rather than required for every image.} It performs perception-guided context acquisition ($\hat c=S(R(Q(z)))$) in three phases: Plan, Search, and Summary.

\textbf{1. Plan ($Q$)}: Using the Stage-I keywords, the MLLM, guided by a prompt designed for image metaphor tasks, generates five search questions aimed at the latent meanings, cultural references, or background information of keyword combinations.

\textbf{2. Search ($R$)}: Each question is routed by a Self-Judge mechanism to one of two knowledge sources.
\begin{enumerate}[label=(\alph*), topsep=0pt, partopsep=0pt, itemsep=2pt, parsep=0pt, leftmargin=*]
    \item \textbf{Self-Judge}: The MLLM outputs a confidence score from 1 to 5 for answering the question from its own knowledge, together with a decision. The rubric lowers the score for internet or meme culture, highly popular knowledge, real-time content, niche entities, and questions involving more than two entities. Questions scored $>3$ go to ModelSearch and questions scored $\le 3$ to WebSearch when the decision agrees; inconsistent or malformed outputs default to ModelSearch. This zero-shot self-assessment follows adaptive retrieval \citep{asai2024selfrag}.
    \item \textbf{ModelSearch}: The MLLM answers the question from its own parametric knowledge with a prompt specialized for explaining memes and image metaphors.
    \item \textbf{WebSearch}: Following the planner--searcher design of MindSearch \citep{chen2024mindsearch}, a planner decomposes the question into sub-questions organized as a dependency graph, and a searcher issues each sub-question to the DuckDuckGo search engine, selects and reads pages, and answers it; the planner then composes the final answer.
\end{enumerate}
Queries are text generated from Stage-I outputs: neither the image nor the benchmark question or options are sent to the search engine, and pages from the benchmark sources are excluded from the retrieved results. This reduces direct leakage, although pages that discuss the same images elsewhere cannot be ruled out.

\textbf{3. Summary ($S$)}: \textbf{RankSummary} ranks the five question--answer pairs by their relevance to the metaphor implied by the image description and keeps the top three; \textbf{RefineSummary} rewrites them, guided by the ranking reasons, into one concise search summary by removing irrelevant or redundant information and reconciling the remaining pieces. This summary is the context $\hat c$ for Stage~III.

\subsection{Stage III: Reasoning}

This stage addresses the \emph{alignment gap} -- how to derive emergent implications from the integration of compositional visual structures and acquired context. The MLLM receives the original image together with the description, the keywords, and the search summary. For MCQ, it articulates its reasoning inside \texttt{<think>}\dots\texttt{</think>} tags, connecting visual cues, keywords, and external knowledge, and gives the option inside \texttt{<answer>} tags; for OSQ, it states the implication concisely under the same instruction given to all baselines. The explicit trace exposes a readable rationale, which we do not assume faithfully reflects the model's internal computation.

Together, the three stages realize a complete test-time contextual alignment pipeline: from extracting compositional structure, through acquiring missing context, to aligning both for emergent implication inference -- all without any parameter updates. Table~\ref{tab:positioning} contrasts this design with prior decompositions, and Algorithm~\ref{alg:pipeline} specifies the complete inference procedure.

\SetAlCapHSkip{0pt}
\SetInd{0.5em}{0.5em}
\SetAlgoNlRelativeSize{-1}
\IncMargin{1em}
\SetKwComment{Comment}{/* }{ */}

\begin{algorithm}[t]
\small
\DontPrintSemicolon
\SetAlgoLined
\caption{Let Androids Dream (LAD)}
\label{alg:pipeline}
\KwIn{Image $x$; task $T\in\{\text{MCQ},\text{OSQ}\}$ (question and options for MCQ)}
\KwOut{Answer $\hat y$}
\BlankLine
\tcp{\small Stage I: Perception, $z=\phi(x)$}
$d \leftarrow \MLLM.\text{Describe}(x)$ \Comment*[r]{\footnotesize one-paragraph description}
$k \leftarrow \MLLM.\text{ExtractKeywords}(d)$ \Comment*[r]{\footnotesize 7 keywords: emotion, domain, rhetoric}
\tcp{\small Stage II: Search, $\hat c=S(R(Q(z)))$}
$\mathcal{Q} \leftarrow \MLLM.\text{PlanQuestions}(k)$ \Comment*[r]{\footnotesize 5 search questions}
$\mathcal{A} \leftarrow \emptyset$\;
\For{each $q \in \mathcal{Q}$}{
    $(s, \mathit{dec}) \leftarrow \MLLM.\text{SelfJudge}(q)$ \Comment*[r]{\footnotesize $s\in\{1,\dots,5\}$}
    \eIf{$s \le 3$ \textbf{and} $\mathit{dec}=\WebSearchStratText$}{
        $a \leftarrow \text{WebSearch}(q)$ \Comment*[r]{\footnotesize external knowledge}
    }{
        $a \leftarrow \MLLM.\text{ModelSearch}(q)$ \Comment*[r]{\footnotesize parametric knowledge}
    }
    $\mathcal{A} \leftarrow \mathcal{A} \cup \{(q, a)\}$\;
}
$(\mathcal{A}_3, r) \leftarrow \MLLM.\text{Rank}(d, \mathcal{A})$ \Comment*[r]{\footnotesize top-3 pairs and ranking reasons}
$\hat c \leftarrow \MLLM.\text{Refine}(r, \mathcal{A}_3)$ \Comment*[r]{\footnotesize context summary}
\tcp{\small Stage III: Reasoning}
$\hat y \leftarrow \MLLM.\text{Reason}(x, d, k, \hat c, T)$\;
\KwRet{$\hat y$}\;
\medskip
\hrule
\medskip
\SetKwProg{Fn}{Function}{}{}
\Fn{\text{WebSearch}{$(q)$}}{
    $G \leftarrow \{q\}$ \Comment*[r]{\footnotesize planner graph, root = question}
    \Repeat{\text{the planner decides to answer}}{
        $\MLLM.\text{Planner}$ adds sub-questions to $G$, each linked to the nodes it depends on\;
        \For{each new node $u$ (in parallel)}{
            $\mathit{res} \leftarrow \text{DuckDuckGo}(\MLLM.\text{RewriteQuery}(u, \text{answers of parents}))$\;
            $\mathit{pages} \leftarrow \text{Fetch}(\MLLM.\text{SelectPages}(\mathit{res}, u))$ \Comment*[r]{\footnotesize benchmark sources excluded}
            $\text{answer}(u) \leftarrow \MLLM.\text{AnswerFromPages}(\mathit{pages}, u)$\;
        }
    }
    \KwRet $\MLLM.\text{Planner.Respond}(G)$\;
}
\end{algorithm}

%% file: sec/experiments.tex
\section{Experiment}
\label{sec:experiments}

\subsection{Baselines}

\begin{table}[t]
\centering
\footnotesize
\renewcommand{\arraystretch}{0.9}
\setlength{\tabcolsep}{4pt}
\caption{Overall results on S100: Multiple-Choice Question (MCQ, accuracy \%) and Open-Style Question (OSQ, 1--5). Models are ordered as open-source (by size), closed-source, and reasoning models. \textbf{Bold}/\underline{underline}: best/second best among the first five models and ours (the last seven were evaluated later; Appendix~\ref{app:more_baselines}); arrows: change over the base model (GPT-4o base: Table~\ref{tab:abl_generalization}(a)).}
\label{tab:main_results}
\begin{adjustbox}{max width=0.82\textwidth}
\begin{tabular}{lcc|cc}
\toprule
\textbf{Model} & \multicolumn{2}{c}{\textbf{Multiple-Choice Question}} & \multicolumn{2}{c}{\textbf{Open-Style Question}}\\
\cmidrule(lr){2-3}\cmidrule(lr){4-5}
 & en & zh & en & zh \\
\midrule
Qwen2.5-VL-7B \citep{qwen25vl} & 46 & 40 & 2.34 & 2.58 \\
Qwen2.5-VL-72B \citep{qwen25vl} & 72 & 56 & 1.56 & 3.12 \\
InternVL3-78B \citep{internvl3} & 70 & 74 & 3.42 & 3.70 \\
GPT-5.4 \citep{gpt54} & 74 & 72 & 3.64 & 3.82 \\
Gemini-3.0-Pro \citep{gemini3} & 76 & \underline{76} & 3.82 & 3.96 \\
Claude-Sonnet-5 \citep{claudesonnet5} & 58 & 54 & 2.86 & 3.26 \\
Qwen3.8-max \citep{qwen38max} & 84 & 82 & 3.80 & 4.00 \\
Gemini-3.1-pro \citep{gemini31pro} & 88 & 80 & 3.06 & 3.26 \\
Grok-4.6 \citep{grok46} & 76 & 84 & 3.56 & 3.14 \\
Claude-Opus-4.7 \citep{claudeopus47} & 64 & 54 & 3.02 & 3.14 \\
Gemini-3.8-flash \citep{gemini38flash} & 78 & 84 & 3.46 & 4.00 \\
GPT-5.6-Sol \citep{gpt56sol} & 78 & 58 & 3.12 & 2.94 \\
\midrule
\multicolumn{5}{c}{\textit{Our Method}} \\
\midrule
GPT-4o-mini \citep{gpt4o} & 44 & 42 & 2.98 & 3.36 \\
\ablationindent + LAD (Stage I + III) & 68~\up & 44~\up & 3.84~\up & 3.58~\up \\
\ablationindent + LAD (Stage I + II + III) & \textcolor{cyan}{74}~\up & \textcolor{cyan}{52}~\up & \textcolor{cyan}{4.02}~\up & \textcolor{cyan}{3.66}~\up \\
\ablationindent Improvement & \textcolor{darkred}{+30} & \textcolor{darkred}{+10} & \textcolor{darkred}{+1.04} & \textcolor{darkred}{+0.30} \\
GPT-4o + LAD (Stage I + II + III) & \underline{\textcolor{cyan}{80}}~\up & \textcolor{cyan}{66}~\up & \underline{\textcolor{cyan}{4.14}}~\up & \underline{\textcolor{cyan}{4.26}}~\up \\
Gemini-3.0-Pro + LAD (Stage I + II + III) & \textbf{\textcolor{cyan}{82}}~\up & \textbf{\textcolor{cyan}{78}}~\up & \textbf{\textcolor{cyan}{4.30}}~\up & \textbf{\textcolor{cyan}{4.46}}~\up \\
\bottomrule
\end{tabular}
\end{adjustbox}
\end{table}

\textbf{Models.}
We compare with open-source MLLMs from 7B to 78B parameters, closed-source models, and reasoning models. \rev{Reflecting the two parts of missing context, Table~\ref{tab:main_results} reports both closed-book LAD (Stage~I + III), which draws only on conventions in the model's weights, and full LAD (Stage~I + II + III), which adaptively invokes web search; we therefore also evaluate} \emph{search-augmented} systems: GPT-4o-mini with OpenAI's native search as Stage II, and Grok-3 and GPT-4o with built-in or Perplexity search (Table~\ref{tab:abl_stage_results}(b)). Closed-book systems use the end-to-end prompts in Figures~\ref{fig:none_mcq}--\ref{fig:none_osq}; model identifiers and decoding parameters are in Appendix~\ref{app:experiment_details}, and earlier baselines in Table~\ref{tab:more_baselines}. \rev{All prompt templates (Appendix~\ref{app:appendix_prompts}) were written once, without tuning on the S100 items, and reused unchanged across all backbones and benchmarks (Appendix~\ref{app:limit}).}

\textbf{Evaluation.}
Our main benchmark is S100, a manually selected 100-image subset of II-Bench \citep{liu2024iibenchimageimplicationunderstanding} and CII-Bench \citep{zhang2024mllmsunderstanddeepimplication} (Table~\ref{tab:dataset_statistics}); the full test sets (1,399 and 800 items) are used in Section~\ref{sec:generalization}. To mitigate MCQ bias \citep{li2024multiplechoicequestionsreallyuseful,zheng2024largelanguagemodelsrobust}, we also evaluate OSQ generation with GPT-4o rubric scoring, and report Wilson intervals and significance tests for MCQ (Appendix~\ref{app:stats}).

\subsection{Multiple-Choice Question}

\begin{figure}[t]
    \centering
	\includegraphics[width=0.68\textwidth]{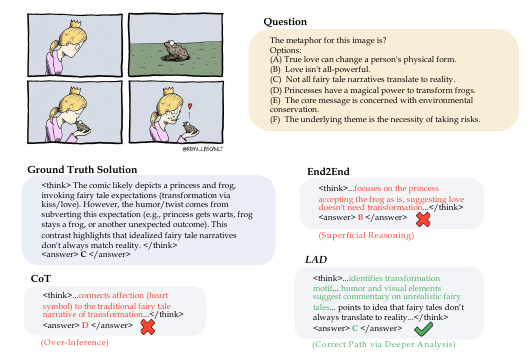}
    \caption{Case study of different methods on Multiple-Choice Question. The \textit{End2End} remains superficial, \textit{CoT} over-infers, while \textit{LAD} gives the correct answer with a context-aligned rationale (outputs abbreviated).}
    \label{fig:case}
\end{figure}

\textbf{Implementation Details.} Each S100 image (comics, posters, illustrations, memes, and traditional artworks) has one six-option question, mostly ``What is the implication in this image?'' (Figure~\ref{fig:case}).

\textbf{Results and Analysis.} Table~\ref{tab:main_results} reports S100 MCQ results. With the lightweight GPT-4o-mini backbone, LAD reaches 74\% on English MCQ -- on par with the closed-book GPT-5.4 and close to Gemini-3.0-Pro (76\%) -- and improves the base model by +30 (en) and +10 (zh) points; with GPT-4o, LAD ties GPT-4o with Perplexity search at 80\% (Table~\ref{tab:abl_stage_results}(b)), and with Gemini-3.0-Pro it reaches 82\%, the best English MCQ accuracy among the directly comparable rows. The seven models evaluated later span 58--88\% (en) and 54--84\% (zh), above GPT-4o-mini + LAD in 12 of 14 model--language pairs; under that relay, an OpenAI-compatible third-party API used for the later evaluations (Appendix~\ref{app:more_baselines}), closed-book GPT-4o scores 52/52 against 74/58 in the original evaluation, so those rows are not directly comparable. The English gain over GPT-4o-mini is significant ($p=0.002$; Appendix~\ref{app:stats}), and both gains hold on the full benchmarks (Section~\ref{sec:generalization}).

\textbf{Qualitative Analysis.} Figure~\ref{fig:case} compares the end-to-end baseline, standard CoT, and LAD on a comic about a princess and a frog. The end-to-end model focuses on the princess accepting the frog and answers that love is not all-powerful (B), staying with the literal scene (\emph{superficial reading}). Standard CoT links the heart symbol to the familiar transformation narrative and answers that princesses can transform frogs (D) (\emph{over-inference}): it reasons explicitly, but with a context recalled from its parameters that does not fit this composition. LAD identifies the transformation motif, reads the humorous twist as a comment on fairy-tale expectations, and selects the correct option that fairy-tale narratives do not always translate to reality (C). The two failure patterns correspond to the perception failure and missing context that Stages~I and~II target (Sections~\ref{sec:perception}--\ref{sec:search}). The superficial-reading failure motivates the emotion, domain, and rhetoric keywords of Stage~I; the II-Bench analysis reports that supplying such labels raises accuracy~\citep{liu2024iibenchimageimplicationunderstanding}.

\subsection{Open-Style Question}
\label{sec:osq}

\begin{figure}[t]
    \centering
    \includegraphics[width=\textwidth]{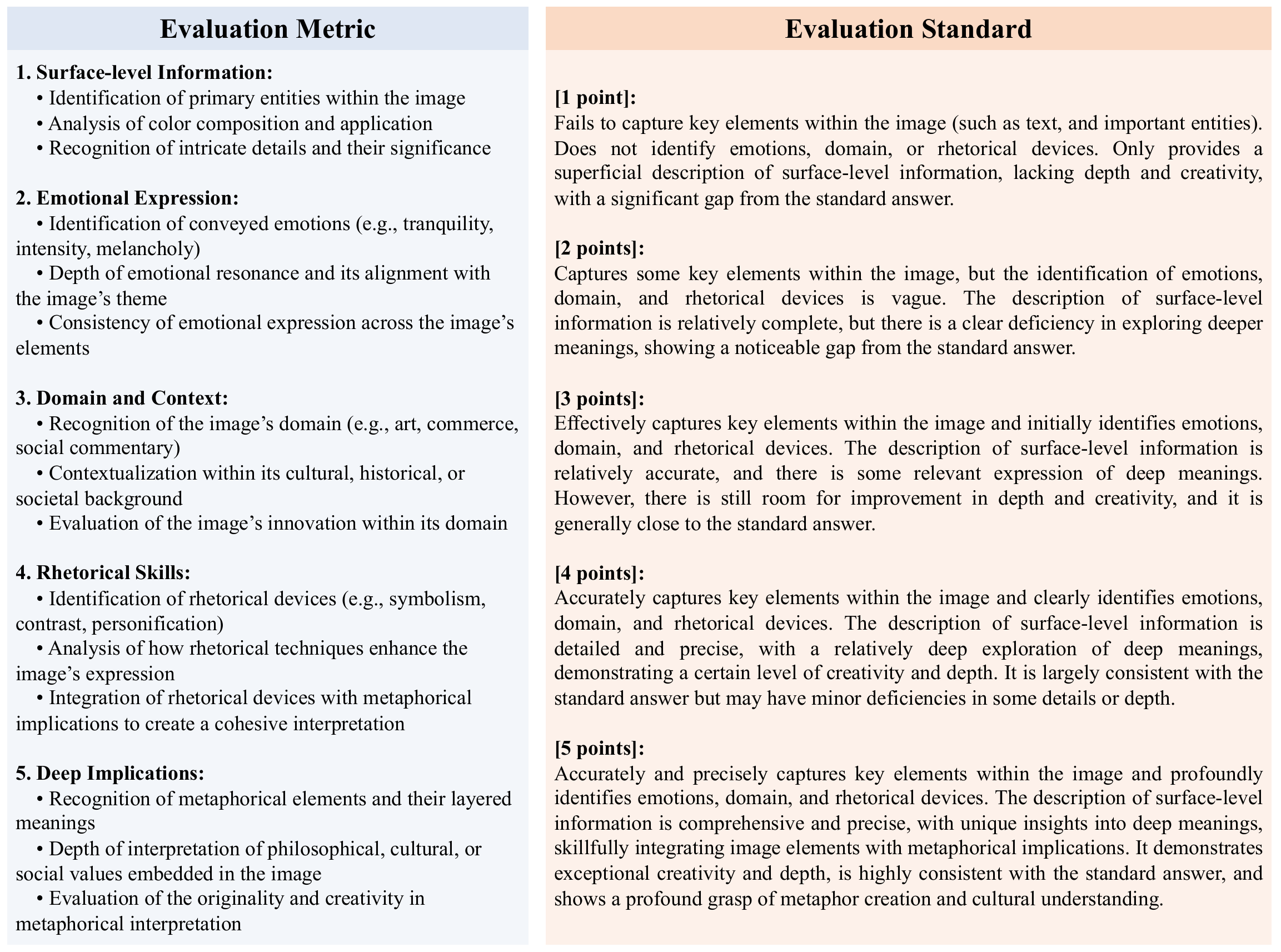}
    \caption{Aspects and level descriptions of the open-style evaluation.}
    \label{fig:osq_evalution}
\end{figure}

\textbf{Evaluation Metric.} OSQ complements MCQ because six-option questions can be answered by eliminating distractors and are sensitive to option order and format~\citep{zheng2024largelanguagemodelsrobust,li2024multiplechoicequestionsreallyuseful,OpenLLMLeaderboard}. Our metric probes both surface-level grounding and the emotion, domain, and rhetoric behind an image through five perspectives (Figure~\ref{fig:osq_evalution}): \textbf{\textit{Surface-level Information}} (whether the answer is grounded in key elements such as embedded text and important entities), \textbf{\textit{Emotional Expression}}, \textbf{\textit{Domain and Context}}, \textbf{\textit{Rhetorical Skills}}, and \textbf{\textit{Deep Implications}}.

\textbf{MLLM-based Automatic Evaluation.} GPT-4o (\texttt{gpt-4o-2024-11-20}, temperature 0) assigns each answer one holistic score from 1 to 5 based on the five perspectives, following the level descriptions in the judge prompt (Appendix~\ref{prompt:judgeosq}), given the image and the human-written reference released with each benchmark. All systems, including LAD, receive the same OSQ instruction (``No need to describe images and text, only answer metaphors \ldots{} try to be concise''); the surface-level aspect concerns whether the answer is grounded in the correct elements, which interpretations typically reference, and applies equally to all systems.

\textbf{Human Consistency Validation.} Sixteen bilingual PhD students and researchers scored model answers for 50 random images (25 per language) with the same rubric; averaging human scores after removing the highest and lowest, the judge is within one point of this average for 96.5\% of English and 94.8\% of Chinese answers (\textbf{95.7\%} overall). \rev{Because this check uses a one-point tolerance on a five-point scale, it validates coarse alignment only, not fine discrimination (Appendix~\ref{app:human_consistency}).}

\textbf{Results and Analysis.} Table~\ref{tab:main_results} reports OSQ results. LAD improves GPT-4o-mini by +1.04 (en) and +0.30 (zh) and GPT-4o by +1.20 and +0.50; Gemini-3.0-Pro with LAD obtains the highest scores in both languages (4.30 and 4.46), followed by GPT-4o with LAD (4.14 and 4.26) and GPT-4o-mini with LAD the second-highest English score (4.02). In contrast, search-augmented baselines show no consistent OSQ improvement (e.g., Perplexity lowers GPT-4o's Chinese score from 3.76 to 3.28). The seven later-evaluated models span 2.86--3.80 in English, below GPT-4o-mini + LAD (4.02) and LAD-CoT (3.84), and reach 4.00 in Chinese vs.\ 3.66 for GPT-4o-mini + LAD; since the later setup also changes GPT-4o's scores, this comparison is indicative only. The Qwen backbones also gain in English (+1.30, +2.06), suggesting that the gains are not limited to the judge's model family; the native-search comparison, which shares Stages I and III with LAD \rev{(4.02 vs.\ 3.62 in English and 3.66 vs.\ 3.34 in Chinese; Table~\ref{tab:abl_stage_results}(b))}, likewise suggests that the rubric does not explain LAD's edge over it. \rev{Under a single judge (\texttt{gpt-4o-2024-11-20}) and this one-point validation tolerance, however, the smaller OSQ gains from adding Stage~II over Stage~I+III (+0.18 in English and +0.08 in Chinese; Table~\ref{tab:main_results}, Figure~\ref{fig:search}(b)) are directional rather than established.}

\subsection{Ablation Study}
\label{sec:ablation}

\begin{table}[t]
\centering
\footnotesize
\renewcommand{\arraystretch}{0.92}
\setlength{\tabcolsep}{3.5pt}
\caption{Stage-wise ablation on S100: (a) reasoning strategies for Stage I+III; (b) search strategies for Stage II. Arrows mark changes relative to the ``w/o'' row of each block; LAD rows are shaded.}
\label{tab:abl_stage_results}
\begin{minipage}[t]{0.40\textwidth}
\centering
\begin{adjustbox}{max width=\linewidth,valign=t}
\begin{tabular}{lcccc}
\toprule
\textbf{Setting} & \multicolumn{2}{c}{\textbf{MCQ}} & \multicolumn{2}{c}{\textbf{OSQ}} \\
\cmidrule(lr){2-3}\cmidrule(lr){4-5}
 & en & zh & en & zh \\
\midrule
\rowcolor{gray!8}\multicolumn{5}{c}{\textit{GPT-4o-mini}} \\
w/o CoT & 44 & 42 & 2.98 & 3.36 \\
Std CoT & 50~\up & 42 & 3.10~\up & 3.28~\down \\
\rowcolor{gray!20} LAD-CoT & \textbf{\textcolor{cyan}{68}}~\up & \textbf{\textcolor{cyan}{44}}~\up & \textbf{\textcolor{cyan}{3.84}}~\up & \textbf{\textcolor{cyan}{3.58}}~\up \\
\bottomrule
\end{tabular}
\end{adjustbox}
\vspace{2pt}
\centerline{\small (a) Reasoning (Stage I+III).}
\end{minipage}
\hfill
\begin{minipage}[t]{0.57\textwidth}
\centering
\begin{adjustbox}{max width=\linewidth,valign=t}
\begin{tabular}{lcccc}
\toprule
\textbf{Setting} & \multicolumn{2}{c}{\textbf{MCQ}} & \multicolumn{2}{c}{\textbf{OSQ}} \\
\cmidrule(lr){2-3}\cmidrule(lr){4-5}
 & en & zh & en & zh \\
\midrule
\rowcolor{gray!8}\multicolumn{5}{c}{\textit{Grok-3}} \\
w/o search & 66 & 64 & 3.24 & \textbf{2.96}\\
Grok built-in search & \textbf{72}~\up & 64 & \textbf{3.25}~\up & 2.92~\down \\
\rowcolor{gray!8}\multicolumn{5}{c}{\textit{GPT-4o}} \\
w/o search & 74 & 58 & \textbf{2.94} & \textbf{3.76}\\
Perplexity search & \textbf{80}~\up & \textbf{66}~\up & 2.88~\down & 3.28~\down \\
\rowcolor{gray!8}\multicolumn{5}{c}{\textit{GPT-4o-mini, with LAD Stages I and III}} \\
w/o search (LAD-CoT) & 68 & 44 & 3.84 & 3.58 \\
OpenAI native search & 72~\up & 48~\up & 3.62~\down & 3.34~\down \\
\rowcolor{gray!20} LAD-Search & \textbf{\textcolor{cyan}{74}}~\up & \textbf{\textcolor{cyan}{52}}~\up & \textbf{\textcolor{cyan}{4.02}}~\up & \textbf{\textcolor{cyan}{3.66}}~\up \\
\bottomrule
\end{tabular}
\end{adjustbox}
\vspace{2pt}
\centerline{\small (b) Search (Stage II).}
\end{minipage}
\end{table}

\begin{table}[t]
\centering
\footnotesize
\renewcommand{\arraystretch}{0.92}
\setlength{\tabcolsep}{5pt}
\caption{\rev{Instructions, stages, and web access separated on one backbone (Qwen2.5-VL-7B, S100 MCQ, accuracy in percent). The single-call row puts the Stage~I and Stage~III metaphor instructions into one prompt with the output cap matched to LAD-CoT. ModelSearch-only keeps every stage, prompt, and decoding setting unchanged but answers each planned Stage-II question without web access. Rows~2--5 are means over three runs; the first and last rows are the single-run values of Table~\ref{tab:abl_generalization}(a).}}
\label{tab:abl_decomp}
\begin{adjustbox}{max width=0.72\linewidth}
\begin{tabular}{lcc}
\toprule
\rev{\textbf{Setting} (Qwen2.5-VL-7B)} & \rev{\textbf{MCQ en} (\%)} & \rev{\textbf{MCQ zh} (\%)} \\
\midrule
\rev{w/o LAD (Table~\ref{tab:abl_generalization}(a))} & \rev{46} & \rev{40} \\
\rev{Std CoT (Figure~\ref{fig:cot_mcq} prompt)} & \rev{46.7} & \rev{42.0} \\
\rev{Single call, LAD metaphor instructions} & \rev{43.3} & \rev{42.0} \\
\rev{LAD-CoT (Stages I+III)} & \rev{54.7} & \rev{44.0} \\
\rev{\ablationindent + Stage II, ModelSearch-only} & \rev{56.7} & \rev{44.7} \\
\rowcolor{gray!20} \ablationindent + Stage II, Self-Judge (full LAD) & \textbf{\textcolor{cyan}{64}} & \textbf{\textcolor{cyan}{46}} \\
\bottomrule
\end{tabular}
\end{adjustbox}
\end{table}

\subsubsection{Stage I (Perception) and Stage III (Reasoning)}

\rev{\textbf{Instructions or stages?} Table~\ref{tab:abl_decomp} separates the metaphor instructions from the staged design on a single backbone. Putting the same emotion, domain, and rhetorical-device instructions into one call, with the output cap matched to LAD-CoT, reaches 43.3\%/42.0\%, which is not above Std CoT (46.7\%/42.0\%). Running the same instructions as separate stages reaches 54.7\%/44.0\%. On this backbone the gain follows the staged decomposition rather than the instruction wording. Stage~II then splits into two parts. Answering its planned questions from the model's own parameters alone adds 2.0 and 0.7 points (56.7\%/44.7\%), while routing to web search reaches 64\%/46\%, so web access supplies 7.3 of the 9.3 English points that Stage~II contributes. The split matches the two parts of missing context in Section~\ref{sec:formulation}: the stages and parametric recall draw out conventions the model already holds, and search adds facts beyond them.}

LAD-CoT (Stage~I+III) improves GPT-4o-mini (MCQ en: 44\%$\rightarrow$68\%, zh: 42\%$\rightarrow$44\%; OSQ en: 2.98$\rightarrow$3.84, zh: 3.36$\rightarrow$3.58), whereas a standard chain-of-thought instruction (Figures~\ref{fig:cot_mcq}--\ref{fig:cot_osq}), which reasons only over the model's parametric knowledge, adds +6 English MCQ points and leaves Chinese MCQ unchanged (Table~\ref{tab:abl_stage_results}(a)). LAD-CoT adds the perceptual abstraction and, for MCQ, \texttt{<think>} reasoning: naming the domain, emotion, and rhetoric of a composition lets the model recall relevant conventions that end-to-end inference misses (Figure~\ref{fig:case}).

\subsubsection{Stage II (Search)}

\begin{figure}[htbp]
    \centering
    \includegraphics[width=\textwidth]{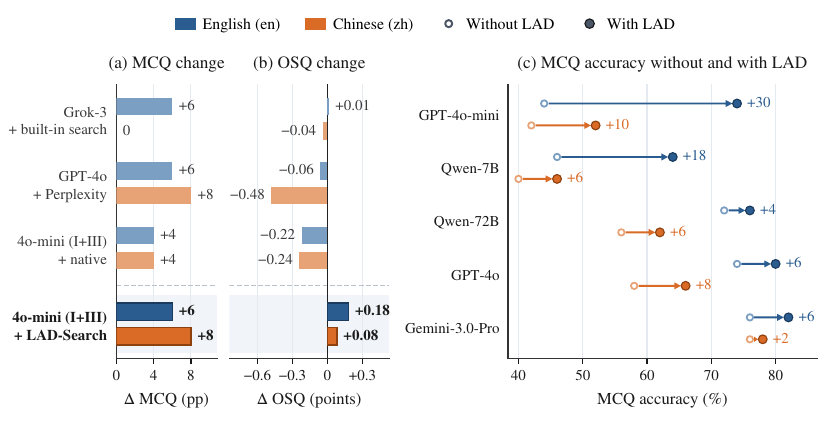}
    \caption{(a, b) Change in MCQ accuracy and OSQ score when search is added, relative to the same system without search (for GPT-4o-mini, relative to LAD-CoT). Other search tools raise MCQ accuracy in five of six cases but change OSQ by $-0.48$ to $+0.01$; LAD-Search raises both. (c) MCQ accuracy of each backbone without LAD (hollow) and with LAD (solid); each arrow shows the improvement (Table~\ref{tab:abl_generalization}(a)).}
    \label{fig:search}
\end{figure}

LAD-Search improves all four metrics over LAD-CoT (MCQ en 68\%$\rightarrow$74\%, zh 44\%$\rightarrow$52\%; OSQ en 3.84$\rightarrow$4.02, zh 3.58$\rightarrow$3.66). In contrast, the other search tools usually raise MCQ accuracy but do not consistently raise OSQ (Table~\ref{tab:abl_stage_results}(b); Figure~\ref{fig:search}): with the backbone and Stages I and III fixed, replacing LAD's Stage II with OpenAI's native search lowers OSQ below the no-search setting (3.84$\rightarrow$3.62, 3.58$\rightarrow$3.34). LAD's search stage, which plans queries from the composition and ranks answers by implication, outperforms both alternatives on all four metrics. \rev{Two controls that would isolate this further, a search-and-answer baseline on the same backbone with the same engine and budget, and separate ablations of question planning and of implication-aware ranking, are left to future work (Appendix~\ref{app:limit}).}

\subsubsection{Different Base Models}

\begin{table}[t]
\centering
\footnotesize
\renewcommand{\arraystretch}{0.92}
\caption{Generalization results: (a) LAD with all three stages run by each backbone; (b) general VQA benchmarks; (c) full benchmarks (II-Bench: 1,399; CII-Bench: 800 items), with approximate 95\% Wilson half-widths for MCQ. \rev{Full LAD uses web search; the Stages I+III row and all other rows are closed-book.} \rev{Full} LAD rows are shaded.}
\label{tab:abl_generalization}
\begin{minipage}[t]{0.38\textwidth}
\centering
\vspace{0pt}
\setlength{\tabcolsep}{3.98pt}
\begin{adjustbox}{max width=\linewidth}
\begin{tabular}{lcccc}
\toprule
\textbf{Setting} & \multicolumn{2}{c}{\textbf{MCQ} (\%)} & \multicolumn{2}{c}{\textbf{OSQ}} \\
\cmidrule(lr){2-3}\cmidrule(lr){4-5}
 & en & zh & en & zh \\
\midrule
\rowcolor{gray!8}\multicolumn{5}{c}{\textit{Qwen2.5-VL-7B}} \\
w/o LAD & 46 & 40 & 2.34 & 2.58 \\
\rowcolor{gray!20} w/ LAD & \textbf{\textcolor{cyan}{64}}~\up & \textbf{\textcolor{cyan}{46}}~\up & \textbf{\textcolor{cyan}{3.64}}~\up & \textbf{\textcolor{cyan}{3.36}}~\up \\
\rowcolor{gray!8}\multicolumn{5}{c}{\textit{GPT-4o-mini}} \\
w/o LAD & 44 & 42 & 2.98 & 3.36 \\
\rowcolor{gray!20} w/ LAD & \textbf{\textcolor{cyan}{74}}~\up & \textbf{\textcolor{cyan}{52}}~\up & \textbf{\textcolor{cyan}{4.02}}~\up & \textbf{\textcolor{cyan}{3.66}}~\up \\
\rowcolor{gray!8}\multicolumn{5}{c}{\textit{Qwen2.5-VL-72B}} \\
w/o LAD & 72 & 56 & 1.56 & 3.12 \\
\rowcolor{gray!20} w/ LAD & \textbf{\textcolor{cyan}{76}}~\up & \textbf{\textcolor{cyan}{62}}~\up & \textbf{\textcolor{cyan}{3.62}}~\up & \textbf{\textcolor{cyan}{3.68}}~\up \\
\rowcolor{gray!8}\multicolumn{5}{c}{\textit{GPT-4o}} \\
w/o LAD & 74 & 58 & 2.94 & 3.76 \\
\rowcolor{gray!20} w/ LAD & \textbf{\textcolor{cyan}{80}}~\up & \textbf{\textcolor{cyan}{66}}~\up & \textbf{\textcolor{cyan}{4.14}}~\up & \textbf{\textcolor{cyan}{4.26}}~\up \\
\rowcolor{gray!8}\multicolumn{5}{c}{\rev{\textit{Gemini-3.0-Pro}}} \\
\rev{w/o LAD} & \rev{76} & \rev{76} & \rev{3.82} & \rev{3.96} \\
\rowcolor{gray!20} w/ LAD & \textbf{\textcolor{cyan}{82}}~\up & \textbf{\textcolor{cyan}{78}}~\up & \textbf{\textcolor{cyan}{4.30}}~\up & \textbf{\textcolor{cyan}{4.46}}~\up \\
\bottomrule
\end{tabular}
\end{adjustbox}
\\[2pt]{\small (a) Different base models (S100).}
\end{minipage}%
\hfill
\begin{minipage}[t]{0.585\textwidth}
\centering
\vspace{0pt}
\begingroup
\setlength{\tabcolsep}{2.2pt}
\begin{adjustbox}{max width=\linewidth}
\begin{tabular}{lccc}
\toprule
\textbf{Model} & \textbf{MMMU} & \textbf{SeedBench} & \textbf{MMStar} \\
\midrule
GPT-4o-mini & 59.4 & 72.8 & 54.8 \\
\rev{\ablationindent + Stages I+III (closed-book)} & \rev{62.4} & \rev{74.5} & \rev{57.8} \\
\rowcolor{gray!20} \ablationindent + LAD (full) & \textcolor{cyan}{67.9} & \textbf{\textcolor{cyan}{77.2}} & \textcolor{cyan}{60.3} \\
GPT-4o & \textbf{70.7} & 76.7 & \textbf{65.1} \\
\bottomrule
\end{tabular}
\end{adjustbox}
\endgroup
\\[2pt]{\small (b) General VQA benchmarks (accuracy, \%).}

\vspace{15.2pt}
\begingroup
\makeatletter
\protected@edef\@currentlabel{\thetable(c)}%
\label{tab:full}%
\makeatother
\setlength{\tabcolsep}{6.34pt}
\begin{adjustbox}{max width=\linewidth}
\begin{tabular}{lcccc}
\toprule
 & \multicolumn{2}{c}{\textbf{MCQ} (\%)} & \multicolumn{2}{c}{\textbf{OSQ}} \\
\cmidrule(lr){2-3}\cmidrule(lr){4-5}
\textbf{Model} & II & CII & II & CII \\
\midrule
GLM-4.1V-8B & 70.0$_{\pm2.4}$ & 46.3$_{\pm3.4}$ & 2.83 & 3.06 \\
GPT-4o & 72.6$_{\pm2.3}$ & 54.1$_{\pm3.4}$ & 3.86 & 4.06 \\
InternVL3-78B & 78.2$_{\pm2.2}$ & 64.0$_{\pm3.3}$ & 3.68 & 4.06 \\
Claude-4.6-Sonnet & 81.0$_{\pm2.1}$ & \textbf{64.5}$_{\pm3.3}$ & 3.62 & 3.82 \\
GPT-4o-mini & 63.5$_{\pm2.5}$ & 35.6$_{\pm3.3}$ & 2.93 & 3.29 \\
\rowcolor{gray!20} LAD (GPT-4o-mini) & \textbf{\textcolor{cyan}{81.2}}$_{\pm2.0}$ & \textcolor{cyan}{53.8}$_{\pm3.4}$ & \textbf{\textcolor{cyan}{4.22}} & \textbf{\textcolor{cyan}{4.31}} \\
\bottomrule
\end{tabular}
\end{adjustbox}
\endgroup
\\[2pt]{\small (c) Full benchmarks (II-Bench: 1,399; CII-Bench: 800 items).}
\end{minipage}
\end{table}

To test whether LAD is model-specific, we apply the full pipeline -- all three stages, including Self-Judge, ModelSearch, and the WebSearch planner and searcher, run by the respective backbone with unchanged prompts -- to the open-source Qwen2.5-VL-7B and 72B, to GPT-4o, \rev{and to Gemini-3.0-Pro} (Table~\ref{tab:abl_generalization}(a)). LAD improves every backbone on every metric. English MCQ gains are larger for the two backbones with low closed-book accuracy (44--46\%: +30 and +18) than for the \rev{three} with high accuracy (\rev{72--76\%: +4, +6 and +6}; Figure~\ref{fig:search}(c)), while Chinese gains \rev{follow the same ceiling pattern: +6 to +10 where closed-book accuracy is 40--58\%, and +2 on Gemini-3.0-Pro at 76\%}. OSQ improves for all backbones (+0.30 to +2.06); the largest gain starts from Qwen2.5-VL-72B's closed-book English score of 1.56 (despite 72\% MCQ accuracy); its cause was not identified.

\subsubsection{Generalization Experiment}
\label{sec:generalization}

\textbf{Experiments on Full Benchmarks.} We evaluate LAD on II-Bench and CII-Bench at full scale (Table~\ref{tab:full}). LAD raises GPT-4o-mini from 63.5\% to 81.2\% and from 35.6\% to 53.8\% ($p<10^{-12}$ for both under approximate unpaired two-proportion $z$-tests; Table~\ref{tab:stats}, Appendix~\ref{app:stats}) and from 2.93 to 4.22 and 3.29 to 4.31 on OSQ, the highest OSQ scores among the compared systems. On II-Bench, it scores above GPT-4o (72.6\%, $p<10^{-7}$) and close to Claude-4.6-Sonnet (81.0\%; difference not significant); on CII-Bench it is close to GPT-4o (54.1\%; not significant) but below InternVL3-78B and Claude-4.6-Sonnet. LAD's gains thus hold at scale; the full-benchmark baselines are closed-book, and search-controlled comparisons exist only on S100 (Table~\ref{tab:abl_stage_results}(b)). \rev{Stage-wise ablations on these two full benchmarks, with paired tests over per-item predictions, are left to future work (Appendix~\ref{app:limit}).}

\textbf{Experiments on General VQA Benchmarks.} To examine whether LAD applies beyond metaphor understanding, we evaluate it on MMMU~\citep{yue2023mmmu} (expert-level multi-discipline reasoning), SeedBench (general understanding), and MMStar (general understanding) with GPT-4o-mini and unchanged prompts (Table~\ref{tab:abl_generalization}(b)). LAD improves GPT-4o-mini on all three (+8.5 on MMMU, +4.4 on SeedBench, +5.5 on MMStar) and exceeds GPT-4o on SeedBench (77.2 vs.\ 76.7). \rev{Closed-book Stages I+III, which add no retrieved text, account for +3.0, +1.7 and +3.0 of this, and Stage~II adds a further +5.5, +2.7 and +2.5. The Stages I+III row is therefore a controlled comparison with the closed-book base model, whereas} the full-LAD row uses web search\rev{, so it is not}: with WebSearch enabled under the same exclusion of benchmark sources, part of the gain, especially on knowledge-intensive MMMU questions, may come from retrieved domain knowledge. We therefore report the full-LAD row as evidence that the pipeline transfers without task-specific changes, not as evidence of a reasoning improvement independent of retrieval.

%% file: sec/discussion.tex
\section{Discussion}
\label{sec:discussion}

\textbf{Perception, search, and reasoning.} Stage~I turns images into compositional abstractions of objects, emotion, domain, and rhetoric, and with Stage~III accounts for most of the \rev{GPT-4o-mini} English MCQ gain\rev{, though not on Qwen2.5-VL-7B, where Stage~II contributes 9.3 of the 18 English points (Table~\ref{tab:abl_decomp})}; Stage~II supplies \rev{additional} cultural and social background, and its OSQ benefit disappears when it is replaced by generic search, consistent with retrieved context helping more when it is selected for the composition. Since entity facts can rule out MCQ distractors without explaining how entities combine, evaluations should also test generation (cognitive analogy: Appendix~\ref{app:detailed_discussion}).

\textbf{Expected failure cases.} Because queries are derived from the Stage~I description, LAD cannot recover context for a decisive element that the description omits, such as small embedded text; off-topic or contradictory search results can still enter the context, since ranking only selects among the retrieved answers; and for abstract artworks whose meaning rests on style rather than recognizable entities, keyword queries retrieve little useful context.

\textbf{Test-time vs.\ train-time contextual alignment.} Contextual alignment can also be internalized by end-to-end reinforcement learning, as in MetaphorStar \citep{zhang2026metaphorstar}. LAD instead needs no parameter updates, applies to frozen backbones, and exposes an inspectable context at the cost of 15--45 seconds of search per image (Appendix~\ref{app:experiment_details}); training a policy to plan questions from $z$ could combine both.

%% file: sec/conclusion.tex
\section{Conclusion}
\label{sec:conclusion}

We reformulate image metaphor understanding as a context-dependent compositional reasoning problem, in which meaning emerges from the interaction of visual elements under cultural and social context rather than from isolated recognition. We identify \emph{missing context} as a failure mode of frozen MLLMs and formalize \emph{Contextual Alignment} as inference over a latent context. Our framework, Let Androids Dream (LAD), realizes contextual alignment with training-free test-time computation through perception, search, and reasoning: it lifts a lightweight GPT-4o-mini from 44\% to 74\% English MCQ accuracy and from 2.98 to 4.02 English OSQ score on S100, improves all \rev{five} backbones we test, and improves open-ended interpretation where generic search does not.
More broadly, many multimodal reasoning tasks require aligning compositional visual structure with knowledge that is not present in the image. For such context-dependent reasoning, how inference-time computation acquires and selects context can matter as much as model scale. Combining test-time context acquisition with train-time alignment is a promising direction for future work.

%% file: sec/appendix.tex
\section*{Appendix Overview}
\begin{itemize}
    \item \ref{app:limit}: Extended limitations and expected failure cases.
    \item \ref{app:experiment_details}: Implementation details (models, decoding, routing, retrieval, baselines, cost).
    \item \ref{app:more_baselines}: Additional baselines.
    \item \ref{app:stats}: Statistical analysis.
    \item \ref{app:human_consistency}: Human evaluation study.
    \item \ref{app:statistic}: S100 benchmark statistics.
    \item \ref{app:detailed_discussion}: Cognitive motivation of the design.
    \item \ref{app:appendix_prompts}: English prompts, verbatim.
\end{itemize}

\section{Extended Limitations and Expected Failure Cases}
\label{app:limit}

\textbf{Evaluation scale.} S100 contains 50 items per language, so a single item changes accuracy by 2 points and the 95\% interval half-width is 11--13 points. We use S100 for the broad model comparison, the ablations, and OSQ, where each system requires one judged answer per image, and we base significance claims on the full II-Bench and CII-Bench (Table~\ref{tab:full}, Appendix~\ref{app:stats}). Ablations on the full benchmarks would give tighter estimates of each component's contribution.

\textbf{Granularity of interventions.} The component analysis adds or replaces whole stages. The formulation of Section~\ref{sec:formulation} suggests finer interventions that we did not run: replacing $\hat c$ with the context of another image (to rule out that any additional text helps), supplying gold annotations of emotion, domain, and rhetoric as $\hat c$ (an upper reference; the II-Bench analysis~\citep{liu2024iibenchimageimplicationunderstanding} reports that such labels raise accuracy), removing the image from Stage III (to measure how much the answer depends on visual input once the description is available), \rev{and routing all questions to WebSearch alone; Table~\ref{tab:abl_decomp} reports the ModelSearch-only side of that comparison. Four further controls are left to future work: a search-and-answer baseline on the same backbone with the same search engine and a comparable budget; closed-book Stages I+III on the full II-Bench and CII-Bench with paired tests; open-style scoring by independent judge models and by blinded human pairwise comparison; and separate ablations inside Stage~II of question planning (composition-guided against entity-only queries) and of implication-aware ranking against unranked selection.}

\textbf{Prompt sensitivity.} The prompt templates were written once and reused without change for Qwen2.5-VL-7B/72B, GPT-4o, Gemini-3.0-Pro, and the general benchmarks (with Chinese versions for Chinese images); this reuse is not evidence of robustness, and we did not measure how results vary under paraphrased instructions.

\textbf{Judge.} OSQ uses one judge model (\texttt{gpt-4o-2024-11-20}; the \texttt{gpt-4o} alias for the later-evaluated models). The judge prompt was written for internet memes and is applied to all image types; its criteria on key elements, emotion, domain, rhetoric, and depth are general, but the persona may be less suited to traditional paintings. Human agreement (Appendix~\ref{app:human_consistency}) and same-backbone comparisons (Section~\ref{sec:osq}) mitigate but do not remove judge-specific preferences.

\textbf{Retrieval dependence and cost.} Results depend on the search engine and its index at run time and can drift. The search stage adds 15--45 seconds and 860--1,070 tokens per image (Appendix~\ref{app:experiment_details}); caching repeated queries and early stopping when Self-Judge is confident would reduce this.

\textbf{Expected failure cases.} By construction, LAD cannot recover from a Stage-I description that omits the decisive element (e.g., small embedded text), because the keywords and queries are derived from it; Stage III still receives the image, which can partly compensate. When retrieved pages are off-topic or contradictory, the ranking step can only choose among the five answers, so irrelevant context may still enter $\hat c$. For highly abstract artworks whose implication rests on style rather than recognizable entities, keyword queries retrieve little useful context. Finally, retrieved internet-culture content can carry stereotypes that the model may reproduce.

\section{Implementation Details}
\label{app:experiment_details}

\textbf{Backbones.} The default backbone is GPT-4o-mini (\texttt{gpt-4o-mini-2024-07-18}). In the backbone study (Table~\ref{tab:abl_generalization}(a)), each backbone runs all three stages, including Self-Judge, ModelSearch, and the WebSearch planner and searcher, with identical prompts. API models are queried through their official APIs; open-source models are served locally.

\textbf{Decoding.} top\_p is 0.9 for all calls. Temperatures: description 0.7, keyword extraction 0.7, question planning 0.7, Self-Judge 0.0, ModelSearch 0.5, ranking 0.0, refinement 0.5, OSQ reasoning 0.7, MCQ reasoning 0.5. Closed-book baselines use temperature 0.5 for MCQ and 0.7 for OSQ with the prompts in Appendix~\ref{app:appendix_prompts}. MCQ answers are parsed from the \texttt{<answer>} tag (LAD) or the ``Answer: \$LETTER'' pattern (baselines).

\textbf{Self-Judge.} The judge outputs ``[score, decision]'' with an integer score from 1 to 5 (prompt in Appendix~\ref{prompt:judge}). A question is sent to WebSearch if the score is at most 3 and the decision is WebSearch, and to ModelSearch if the score is above 3 and the decision is ModelSearch; inconsistent or malformed outputs default to ModelSearch.

\textbf{WebSearch.} The planner--searcher follows MindSearch~\citep{chen2024mindsearch}, using the same backbone as LAD for planning, query rewriting, page selection, and answering, and DuckDuckGo as the search engine. Pages from the benchmark sources (the II-Bench and CII-Bench project, dataset, and paper pages) were excluded from the results in the search configuration of our runs; this filter is not part of the released MindSearch client code. Queries are text only; the image and the benchmark question or options are never sent to the search engine.

\textbf{Search-augmented baselines.} ``Grok-3 + built-in search'' enables Grok-3's native web search; ``GPT-4o + Perplexity'' combines GPT-4o with Perplexity search; ``GPT-4o-mini + native search'' keeps LAD's Stage I and Stage III and replaces Stage II (planning, routing, retrieval, ranking, and condensing) with OpenAI's native web search, so the backbone and the Stage I/III prompts are shared with LAD. Because the search engine and the selection of retrieved content both differ, the comparison tests the acquisition pipeline as a whole rather than a single component.

\textbf{OSQ references and judge.} The reference answer is the human-written explanation released with II-Bench (English) and the concatenation of the human-written metaphorical meaning and explanation released with CII-Bench (Chinese). The judge is GPT-4o (\texttt{gpt-4o-2024-11-20}) at temperature 0; it receives the image, the reference, and the answer, and each answer is scored once.

\textbf{Cost.} Measured with the search questions executed concurrently, each search question takes 4--9 seconds and the whole search stage (including planning, ranking, and refinement) takes 15--45 seconds and consumes roughly 860--1,070 tokens per image. Stages I and III add three MLLM calls (description, keywords, reasoning).

\section{Additional Baselines}
\label{app:more_baselines}

Table~\ref{tab:more_baselines} lists the earlier S100 baselines kept separate from Table~\ref{tab:main_results} for readability, including the search-augmented Grok-3 and GPT-4o systems also shown in Table~\ref{tab:abl_stage_results}(b); GPT-4o and Grok-3 are the base models of these systems and of the GPT-4o LAD rows. The later-evaluated models in Tables~\ref{tab:main_results} and~\ref{tab:more_baselines} (Gemini-3.8-flash, Qwen3.8-max, Claude-Sonnet-5, GPT-5.6-Sol, Grok-4.6, Gemini-3.1-pro, Qwen3-VL-Plus in thinking mode, Claude-Opus-4.7) were evaluated in September 2026 through an OpenAI-compatible API relay with the end-to-end prompts of Figures~\ref{fig:none_mcq}--\ref{fig:none_osq} (MCQ temperature 0.5, OSQ 0.7, one sample) and scored by the \texttt{gpt-4o} alias with the same judge prompt, because \texttt{gpt-4o-2024-11-20} was not available. Gemini-3.1-pro, Qwen3-VL-Plus, and Claude-Opus-4.7 are the \texttt{gemini-3.1-pro-preview}, \texttt{qwen3-vl-plus-think}, and \texttt{claude-opus-4-7} endpoints; Grok-4.6, Qwen3.8-max, and Qwen3-VL-Plus have 49 scored Chinese items (accuracy over 49) because one request failed repeatedly; refusals count as incorrect. We cannot verify which model versions the relay serves, and the cause of the GPT-4o differences was not identified. For reasoning endpoints, the relay returns only the final answer, occasionally preceded by a few words of reasoning residue, which we did not remove. Each endpoint passed a check in which it correctly described the text and costumes of an S100 comic. Under this setup, closed-book GPT-4o obtains 52/52 (MCQ) and 2.88/3.06 (OSQ), versus 74/58 and 2.94/3.76 in the original evaluation, so these rows are not directly comparable with the other rows.

\begin{table}[h]
\centering
\small
\caption{Additional baselines on S100 (MCQ accuracy \%, OSQ 1--5; Qwen3-VL-Plus was evaluated in the later setup).}
\label{tab:more_baselines}
\begin{tabular}{lcccc}
\toprule
\textbf{Model} & MCQ en & MCQ zh & OSQ en & OSQ zh \\
\midrule
GPT-4o~\citep{gpt4o} & 74 & 58 & 2.94 & 3.76 \\
GLM-4.1V-8B~\citep{glm4v} & 60 & 52 & 2.60 & 2.96 \\
Grok-3~\citep{grok3} & 66 & 64 & 3.24 & 2.96 \\
DeepSeek-VL2~\citep{wu2024deepseekvl2mixtureofexpertsvisionlanguagemodels} & 46 & 36 & 2.82 & 2.86 \\
GLM-4V-plus~\citep{glm4v} & 64 & 64 & 3.01 & 3.12 \\
Claude-4-Sonnet~\citep{claude4sonnet} & 60 & 62 & 3.46 & 3.86 \\
QVQ-72B~\citep{qvq-72b-preview} & 62 & 56 & 3.10 & 3.42 \\
Doubao-1.5-thinking-vision-pro~\citep{doubao-1.5-thinking-vision-pro} & 66 & 66 & 3.16 & 3.90 \\
Grok-3-reasoning~\citep{grok3} & 74 & 64 & 3.06 & 2.92 \\
Claude-4.6-Sonnet~\citep{claude46sonnet} & 64 & 62 & 3.56 & 3.72 \\
Claude-4.6-Opus~\citep{claude46opus} & 70 & 82 & 3.60 & 3.94 \\
\midrule
Grok-3 + built-in search & 72 & 64 & 3.25 & 2.92 \\
GPT-4o + Perplexity search & 80 & 66 & 2.88 & 3.28 \\
\midrule
Qwen3-VL-Plus (thinking)~\citep{qwen3vl} & 88 & 76 & 3.08 & 4.02 \\
\bottomrule
\end{tabular}
\end{table}

\section{Statistical Analysis}
\label{app:stats}

For an accuracy $\hat p$ on $n$ items (Table~\ref{tab:full}) we report the 95\% Wilson interval
\begin{equation*}
\Big(\hat p+\tfrac{z^2}{2n}\pm z\sqrt{\tfrac{\hat p(1-\hat p)}{n}+\tfrac{z^2}{4n^2}}\Big)\Big/\Big(1+\tfrac{z^2}{n}\Big),\qquad z=1.96.
\end{equation*} Differences between two systems are approximated with an unpaired two-proportion $z$-test on correct-answer counts reconstructed from the reported accuracies; a paired McNemar test would require per-item predictions. The test ignores that both systems answer the same items; it is conservative when correctness is positively correlated across systems, which is typical but could not be verified without per-item predictions, so the $p$-values should be read as approximate. Table~\ref{tab:stats} lists the comparisons used in the paper.

\begin{table}[h]
\centering
\small
\caption{Significance of key comparisons (two-proportion $z$-test).}
\label{tab:stats}
\setlength{\tabcolsep}{4pt}
\begin{adjustbox}{max width=\textwidth}
\begin{tabular}{llccc}
\toprule
\textbf{Benchmark} & \textbf{Comparison} & \textbf{Accuracies (\%)} & $z$ & $p$ \\
\midrule
II-Bench ($n$=1,399) & GPT-4o-mini + LAD vs.\ GPT-4o-mini & 81.2 vs.\ 63.5 & 10.48 & $1.1\times10^{-25}$ \\
CII-Bench ($n$=800) & GPT-4o-mini + LAD vs.\ GPT-4o-mini & 53.8 vs.\ 35.6 & 7.29 & $3.1\times10^{-13}$ \\
II-Bench & GPT-4o-mini + LAD vs.\ GPT-4o & 81.2 vs.\ 72.6 & 5.38 & $7.3\times10^{-8}$ \\
II-Bench & GPT-4o-mini + LAD vs.\ Claude-4.6-Sonnet & 81.2 vs.\ 81.0 & 0.14 & 0.89 \\
CII-Bench & GPT-4o-mini + LAD vs.\ GPT-4o & 53.8 vs.\ 54.1 & $-$0.15 & 0.88 \\
S100-en ($n$=50) & GPT-4o-mini + LAD vs.\ GPT-4o-mini & 74 vs.\ 44 & 3.05 & 0.002 \\
S100-zh ($n$=50) & GPT-4o-mini + LAD vs.\ GPT-4o-mini & 52 vs.\ 42 & 1.00 & 0.32 \\
S100-en & GPT-4o-mini + LAD vs.\ Gemini-3.0-Pro & 74 vs.\ 76 & $-$0.23 & 0.82 \\
\bottomrule
\end{tabular}
\end{adjustbox}
\end{table}

\section{Human Evaluation Study}
\label{app:human_consistency}

\textbf{Participants.} Sixteen PhD students and researchers, recruited as volunteers within the authors' institutions, took part. All are proficient in English and Chinese and familiar with comics, memes, and visual metaphors. Participation was voluntary and uncompensated, participants were informed about the purpose of the study and how their ratings would be used, and no personal data were collected. The study involved only rating model outputs for public images; no independent institutional ethics review or exemption determination was obtained.

\textbf{Materials and instructions.} We randomly sampled 25 English and 25 Chinese S100 images. For each image, participants saw the image, the human reference answer, and anonymized model answers in random order, and were instructed: ``Score each answer from 1 to 5 using the scoring standard. Judge how well the answer captures the implied meaning of the image with respect to the key elements, emotion, domain, rhetorical devices, and depth of implication, compared with the reference answer. Do not let the length of the answer influence your score.'' The scoring standard was the same level description given to the judge (Appendix~\ref{prompt:judgeosq}).

\textbf{Agreement measure.} Let $h_{i,1},\dots,h_{i,m}$ be the human scores for answer $i$ and $\tilde h_i$ their mean after removing the highest and the lowest score. With judge score $g_i$, agreement is the fraction of answers with $|\tilde h_i-g_i|\le 1$. Trimming removes individual outliers before averaging. \rev{Because $|\tilde h_i-g_i|\le 1$ uses a one-point tolerance on a 5-point scale, this check validates coarse alignment only, not fine discrimination between close system averages such as the +0.08 to +0.18 OSQ differences from adding search in Section~\ref{sec:osq}.} Agreement is 96.5\% for English, 94.8\% for Chinese, and 95.7\% overall. \rev{Scoring by independent judge models and by blinded human pairwise comparison is left to future work.}

\section{S100 Benchmark Statistics}
\label{app:statistic}

S100 was built by selecting 50 images from II-Bench and 50 from CII-Bench that are diverse in domain and image type and rich in visual and cultural content. Table~\ref{tab:dataset_statistics} lists the distribution. Each image has one six-option MCQ; most questions ask for the overall implication, and some ask about specific details that are necessary for it.

\begin{table}[h]
    \centering
    \footnotesize
    \caption{Distribution of domains and image types in S100.}
    \label{tab:dataset_statistics}
    \begin{minipage}[t]{0.45\textwidth}
        \centering
        \begin{tabular}{@{}ll@{}}
            \toprule
            \multicolumn{2}{l}{\textbf{English images (II-Bench)}} \\ \cmidrule(r){1-2}
            Society & 21 (42\%) \\
            Life & 16 (32\%) \\
            Art & 6 (12\%) \\
            Psychology & 4 (8\%) \\
            Others & 3 (6\%) \\
            \midrule
            Multi-panel comic & 16 (32\%) \\
            Single-panel comic & 9 (18\%) \\
            Illustration & 5 (10\%) \\
            Meme & 5 (10\%) \\
            Poster & 5 (10\%) \\
            Painting & 5 (10\%) \\
            Logo & 5 (10\%) \\
            \bottomrule
        \end{tabular}
    \end{minipage}
    \hfill
    \begin{minipage}[t]{0.45\textwidth}
        \centering
        \begin{tabular}{@{}ll@{}}
            \toprule
            \multicolumn{2}{l}{\textbf{Chinese images (CII-Bench)}} \\ \cmidrule(r){1-2}
            Life & 13 (26\%) \\
            Art & 13 (26\%) \\
            Society & 12 (24\%) \\
            Chinese traditional culture & 6 (12\%) \\
            Environment & 5 (10\%) \\
            Politics & 1 (2\%) \\
            \midrule
            Illustration & 15 (30\%) \\
            Single-panel comic & 10 (20\%) \\
            Poster & 8 (16\%) \\
            Meme & 8 (16\%) \\
            Painting & 6 (12\%) \\
            Multi-panel comic & 3 (6\%) \\
            \bottomrule
        \end{tabular}
    \end{minipage}
\end{table}

\section{Cognitive Motivation}
\label{app:detailed_discussion}

The title and epigraph pose two questions. \emph{Do androids dream?}---can a model interpret implications at all---is addressed by Perception and Reasoning, which turn pixels into a structured reading and connect it to a conclusion. \emph{Do they dream of electric sheep?}---are the interpretations specific and grounded in the right context---is addressed by Search, which supplies the cultural and social background the image presupposes.

\textbf{Cognitive motivation.} LAD's design is an engineering analogy, not a model of human cognition. Perception follows a coarse-to-fine order that mirrors how people first form a holistic impression of an image and then identify its key elements; in dual-process terms~\citep{Evans2003InTM}, the description corresponds to the fast holistic impression and the keyword extraction to deliberate analysis. Search follows accounts of active information seeking~\citep{Informationseeking,Wilson2009ActivityTA}: when people meet an unfamiliar reference, they look it up. Reasoning connects the elements with the retrieved background, which resembles the social inference involved in interpreting intent~\citep{kim2025hypothesisdriventheoryofmindreasoninglarge}. We use these analogies only to motivate the decomposition; the experiments evaluate it on its own terms.

\section{Prompts}
\label{app:appendix_prompts}

All prompts are shared by all backbones. We list the English LAD prompts verbatim; \texttt{\{<slot>\}} marks where inputs (description, keywords, questions, retrieved answers, or the MCQ) are inserted. The Chinese prompts are translations with the same structure and are included in the released code, together with the WebSearch planner and searcher prompts from MindSearch.

\lstset{basicstyle=\ttfamily\scriptsize, breaklines=true, breakindent=0pt, columns=fullflexible, frame=single, framesep=3pt, xleftmargin=3pt, xrightmargin=3pt, keepspaces=true, showstringspaces=false}

\input{sec/prompts}

\begin{figure}[h]
    \centering
    \includegraphics[width=0.9\textwidth]{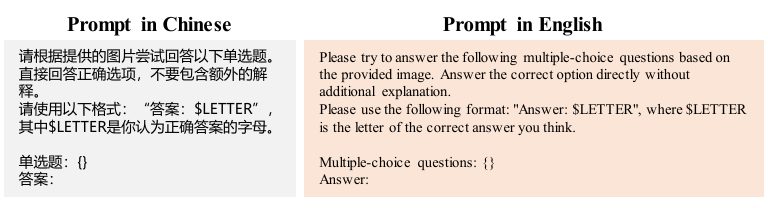}
    \caption{End-to-end baseline prompt for MCQ.}
    \label{fig:none_mcq}
\end{figure}

\begin{figure}[h]
    \centering
    \includegraphics[width=0.9\textwidth]{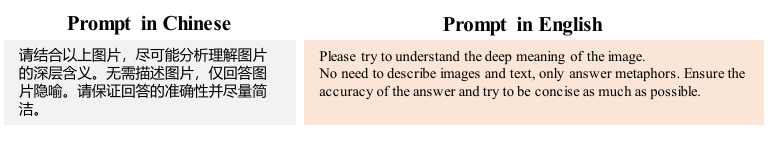}
    \caption{End-to-end baseline prompt for OSQ.}
    \label{fig:none_osq}
\end{figure}

\begin{figure}[h]
    \centering
    \includegraphics[width=0.9\textwidth]{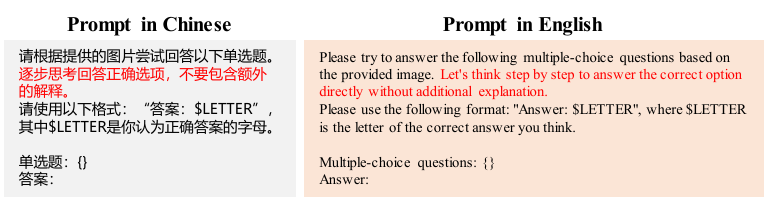}
    \caption{Standard chain-of-thought baseline prompt for MCQ (the ``Std CoT'' row of Table~\ref{tab:abl_stage_results}(a)).}
    \label{fig:cot_mcq}
\end{figure}

\begin{figure}[h]
    \centering
    \includegraphics[width=0.9\textwidth]{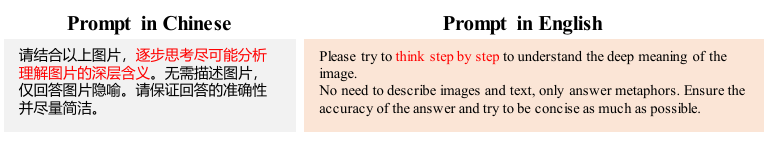}
    \caption{Standard chain-of-thought baseline prompt for OSQ.}
    \label{fig:cot_osq}
\end{figure}

\paragraph{Closed-book baselines.} Figures~\ref{fig:none_mcq}--\ref{fig:none_osq} show the end-to-end prompts used for all closed-book systems, and Figures~\ref{fig:cot_mcq}--\ref{fig:cot_osq} the standard chain-of-thought prompts of Table~\ref{tab:abl_stage_results}(a) (Chinese and English). In the Stage-III prompts, the slot labelled ``Keywords description'' receives the search summary $\hat c$; in the no-search setting (LAD-CoT) no search summary is provided.


%% file: sec/prompts.tex
\par\medskip\noindent\textbf{Stage I: image description.}\label{prompt:desc}\par\smallskip
\begin{lstlisting}
Please provide a description of the image in a paragraph. You should consider the role, text, color, and layout of the image. Focus on the text and important elements in the image. Try to be concise while ensuring the correctness of the description.
\end{lstlisting}

\par\medskip\noindent\textbf{Stage I: keyword extraction.}\label{prompt:kw}\par\smallskip
\begin{lstlisting}
Please provide seven keywords most related to the metaphor based on the image description. You should consider the emotion, domain, and rhetoric in the image description. Focus on the image’s text and important entities in the image description. Note that there may be homophonic memes and puns, distinguish and provide them but do not repeat the same element. Take a deep breath and think step by step, only output the keywords.
Image description: {<slot>}
\end{lstlisting}

\par\medskip\noindent\textbf{Stage II: search-question planning.}\label{prompt:plan}\par\smallskip
\begin{lstlisting}
# Role
You are a researcher familiar with internet culture and memes, skilled at uncovering and analyzing the deeper meanings of internet memes.
## Attention
Your responsibility is to design precise search questions based on keywords that can accurately retrieve meme culture or image metaphors.

## Skills
### Skill 1: Internet Search Skills
- Analyze the provided keywords and select the most relevant ones, understanding their potential meanings and applications in metaphorical culture

### Skill 2: Cultural Analysis
- Deeply explore the background and history of keywords in internet culture, through information organization and analysis, to uncover the deeper meanings behind the keywords

### Skill 3: Metaphor Comprehension
- Understand and analyze the metaphorical meanings of keywords in images, to better comprehend the cultural background of meme images

## Workflow:
1. Analyze the provided keywords and select the most relevant ones, understanding their potential meanings and applications in metaphorical culture.
2. Design specific and targeted search questions to improve the accuracy of the search.
3. Ensure that the search questions can guide users to find cultural backgrounds or metaphorical explanations related to the meme images.

## Constraints
- Search questions should focus on meme culture or image metaphors, avoiding overly broad or irrelevant search results
- Search questions should consider keyword combinations, avoiding single keyword searches as much as possible
- The total number of search questions should be 5, please design the most relevant questions

## Example:
Input:
"""
Keywords: irony, environmental protection, waste, contrast, empty slogans, lack of action, inconsistency between words and deeds
"""
Output:
"""
1. "What are the specific manifestations of the inconsistency between environmental protection slogans and actual actions? "

2. "What is the profound connotation of the popular Internet pictures comparing environmental protection and waste? "

3. "Dual information pictures satirizing environmental protection slogans and lack of action? "

4. "Ironic contrast pictures of the contradiction between slogans and actual actions in environmental protection themes? "

5. "How do Internet pictures show the ironic contrast between environmental protection slogans and wasteful behavior? "
"""

## Solve:
keywords: {<slot>}
\end{lstlisting}

\par\medskip\noindent\textbf{Stage II: Self-Judge routing.}\label{prompt:judge}\par\smallskip
\begin{lstlisting}
# Task
Please evaluate your knowledge level, determine whether the question is suitable for direct answering or requires external knowledge support, and provide a confidence score according to the Evaluation Standard.

## Evaluation Standard
- If the question contains Internet culture and meme culture, the confidence score is less than 3
- If the knowledge contained in the question is highly popular, the confidence score is less than 3
- If the real-time content contained in the question accounts for a large proportion, the confidence score is less than 3
- If the entity contained in the question is relatively niche, the confidence score is less than 3
- If the question contains more than 2 entities, the confidence score is less than 3

## Workflow:
1. <Implicitly> Analyze the content of the question and identify whether it contains Internet culture and meme culture elements.
2. <Implicitly> Evaluate the popularity of the question and the proportion of real-time content.
3. According to the scoring criteria, <Explicitly> provide a confidence score and decision.

## Constraints
- The confidence score must be between 1 and 5. If the score is bigger than 3, select ModelSearch. If the score is smaller than or equal to 3, select WebSearch.

## OutputFormat:
[Confidence score (1-5 points), Decision (ModelSearch/WebSearch)]

## Solve:
Question: {<slot>}
\end{lstlisting}

\par\medskip\noindent\textbf{Stage II: ModelSearch.}\label{prompt:modelsearch}\par\smallskip
\begin{lstlisting}
# Role
You are a researcher familiar with internet culture and memes, skilled at uncovering and analyzing the deeper meanings of internet memes.
## Attention
You are responsible for providing clear, professional and comprehensive explanations of Internet memes and image metaphors based on Internet culture and meme culture, helping users understand the cultural meaning and social impact behind them.

## Skills
### Skill 1: Metaphor Understanding
- Ability to accurately identify the memes or metaphors involved in the picture based on picture analysis, and be able to extract the deep meaning

### Skill 2: Internet Culture Analysis
- Ability to analyze the origin, development and current popularity of memes or image metaphors, so that users can better understand the cultural background of memes

### Skill 3: Providing Explanations
- Ability to provide users with clear and professional explanations of the cultural meaning and social impact behind memes or image metaphors
- Ability to provide structured and easy-to-understand answers

## Workflow:
1. Identify the meme or picture metaphor involved in the question raised by the user.
2. Analyze the origin, development and current popularity of the meme or picture metaphor.
3. Explain the cultural meaning and social impact behind the meme or picture metaphor.
4. Provide a structured and easy-to-understand answer.

## Constraints
- The answer content needs to be logically clear and well-structured to ensure that readers can understand it easily
- The answer content needs to focus on the meme culture or picture metaphor
- The answer part needs to be comprehensive and complete, and there should be no vague expressions such as "based on the above content" to ensure the credibility of the information
- The language style needs to be professional and rigorous, and avoid colloquial expressions
- Maintain uniform grammar and vocabulary to ensure the consistency and coherence of the overall document

## Solve:
Question: {<slot>}
\end{lstlisting}

\par\medskip\noindent\textbf{Stage II: relevance ranking.}\label{prompt:rank}\par\smallskip
\begin{lstlisting}
# Role
You are a researcher familiar with Internet culture and memes, and are good at extracting deep meanings from images, and matching and sorting these meanings with relevant questions and summaries.
## Attention
You are responsible for sorting the relevance of multiple question-summary pairs based on the image description, and selecting the three most relevant ones to help users understand the deep meaning of the image.

## Skills
### Skill 1: Image Analysis
- Ability to conduct a comprehensive analysis based on the user's description of the image

### Skill 2: Metaphor Understanding
- Ability to accurately identify metaphors in images based on image analysis and extract deep meaning

### Skill 3: Relevance Sorting
- Ability to sort the relevance of question-summary pairs based on the metaphorical understanding of the image, based on whether the content of the question-summary pair is related to the image metaphor
- Ability to provide reasons for the corresponding relevance sorting, and analyze the low-relevance content of the question-summary pair

## Workflow:
1. Carefully analyze the image description and <implicitly> extract the metaphors and key elements in the image.
2. Analyze each question-summary pair and <implicitly> evaluate its relevance to the image metaphor.
3. According to the relevance, <implicitly> sort the question-summary pairs.
4. <display> Output the top 3 most relevant question-summary pairs after sorting and the reasons for sorting.

## Constraints
- Sorting should be based on the relevance of the image metaphor to ensure the accuracy and logic of the sorting results
- If a part of the question-summary pair is not related to the image metaphor, please provide it in the corresponding reason
- Only output the top 3 most relevant question-summary pairs after sorting and the reasons for sorting, avoid using markdown format

## Examples:
Input:
"""
Image Description: "A person standing at a crossroads", Question-Summary Pair: "1. Question: "The importance of life choices"; 2. Question: "Decision-making process in the face of difficulties"; 3. Question: "Complexity of urban planning"; 4. Question: "Navigation skills in travel"; 5. Question: "Personal growth and self-discovery""
"""
Output:
"""
1. Question: The importance of life choices
  - Sorting Reason: xxx
2. Question: Decision-making process in the face of difficulties
  - Sorting Reason: xxx
3. Question: Personal growth and self-discovery
  - Sorting Reason: xxx
"""

## Solve:
Image Description: {<slot>}
Question-Summary Pair: {<slot>}
\end{lstlisting}

\par\medskip\noindent\textbf{Stage II: refinement into the context summary.}\label{prompt:refine}\par\smallskip
\begin{lstlisting}
# Role
You are a researcher familiar with Internet culture and memes, and are good at rewriting the selected "question-summary pairs" according to the relevance sorting reasons to improve the appeal of the content and the accuracy of the information.
## Attention
You are responsible for rewriting the selected "question-summary pairs" according to the relevance ranking reasons, removing repeated semantics, parts with low relevance to Internet culture and meme culture, and removing markdown and citation formatting symbols, and finally outputting a fused and rewritten content.
## Skills
### Skill 1: Internet Culture Insight
- Able to deeply understand classic film and television characters, classic film and television plots and meme culture, and accurately grasp network hot spots and popular trends

### Skill 2: Content Analysis and Rewriting
- Able to conduct detailed analysis of "question-summary pairs" and identify the relevance to Internet culture and meme culture
- Able to creatively rewrite content based on analysis results to improve the accuracy and attractiveness of information

### Skill 3: Format Recognition and Processing
- Able to identify and remove markdown and reference format symbols to ensure the simplicity and readability of output content
- Able to integrate rewritten content and connect it with appropriate transition sentences to form a coherent and fluent article

## Workflow:
1. Read and understand the relevance sorting results and all "question-summary pairs" provided by the user.
2. Based on the selection reasons, analyze the relevance of each "question-summary pair" to Internet culture and meme culture.
3. Rewrite the selected 3 "question-summary pairs", remove repeated semantics, low relevance parts, and format symbols.
4. Integrate the rewritten content, connect it with appropriate transition sentences to form a coherent and fluent article.

## Constraints
- The rewritten content should maintain the original meaning and should not change the original information and viewpoints
- The language should be concise, clear, easy to understand, and related to the image metaphor
- Do not use any form of markdown and reference format symbols

## Solve:
Relevance Sorting Results: {<slot>}
Most Relevant Question-Summary Pair: {<slot>}
\end{lstlisting}

\par\medskip\noindent\textbf{Stage III: reasoning for OSQ.}\label{prompt:osq}\par\smallskip
\begin{lstlisting}
Image description: {<slot>}; 
Keywords: {<slot>}; 
Keywords description: {<slot>};
Please combine the image, keywords, and description information and try to understand the deep meaning of the combination of the image and text. No need to describe images and text, only answer metaphors. Ensure the accuracy of the answer and try to be concise as much as possible.
\end{lstlisting}

\par\medskip\noindent\textbf{Stage III: reasoning for MCQ (the MCQ is inserted as ``Question: ... Options: A. ... F. ...'').}\label{prompt:mcq}\par\smallskip
\begin{lstlisting}
Image description: {<slot>}; 
Keywords: {<slot>}; 
Keywords description: {<slot>};
Please combine the image, keywords, and description information and try to understand the deep meaning of the combination of the image and text. Answer the following multiple-choice questions with the correct option $LETTER directly. Output the thinking process in <think> </think> and final correct answer in <answer> </answer> tags. The output format should be as follows: <think>...</think> <answer>...</answer> 
Multiple-choice questions: {<slot>}
Answer:
\end{lstlisting}

\par\medskip\noindent\textbf{OSQ judge (GPT-4o), reproduced as used, including the typo ``Standrad''.}\label{prompt:judgeosq}\par\smallskip
\begin{lstlisting}
# Role
You are an impartial judge who is familiar with Internet culture and memes, and is good at digging out and analyzing the deep meaning of Internet memes.

## Attention
You are responsible for evaluating the quality of the answer provided by the model for Internet culture and memes. Your evaluation should refer to the human answer and image, and score based on the Evaluation Standard.

## Evaluation Standard:
[1 point]: Fails to capture key elements within the image (such as text, and important entities). Does not identify emotions, domain, or rhetorical devices. Only provides a superficial description of surface-level information, lacking depth and creativity, with a significant gap from the standard answer.

[2 points]: Captures some key elements within the image, but the identification of emotions, domain, and rhetorical devices is vague. The description of surface-level information is relatively complete, but there is a clear deficiency in exploring deeper meanings, showing a noticeable gap from the standard answer.

[3 points]: Effectively captures key elements within the image and initially identifies emotions, domain, and rhetorical devices. The description of surface-level information is relatively accurate, and there is some relevant expression of deep meanings. However, there is still room for improvement in depth and creativity, and it is generally close to the standard answer.

[4 points]: Accurately captures key elements within the image and clearly identifies emotions, domain, and rhetorical devices. The description of surface-level information is detailed and precise, with a relatively deep exploration of deep meanings, demonstrating a certain level of creativity and depth. It is largely consistent with the standard answer but may have minor deficiencies in some details or depth.

[5 points]: Accurately and precisely captures key elements within the image and profoundly identifies emotions, domain, and rhetorical devices. The description of surface-level information is comprehensive and precise, with unique insights into deep meanings, skillfully integrating image elements with metaphorical implications. It demonstrates exceptional creativity and depth, is highly consistent with the standard answer, and shows a profound grasp of metaphor creation and cultural understanding.

## Standrad Answer:
Human answer: {<slot>}

## Constraints
- Avoid any position biases and be as objective as possible
- Do not allow the length of the descriptions to influence your evaluation
- Output your final directly by strictly following this format: "[ratings]"

## Solve:
Model answer: {<slot>}
\end{lstlisting}